\documentclass[conference]{IEEEtran}
\usepackage{graphicx} %

\usepackage{cite}
\usepackage[utf8]{inputenc} %
\usepackage[T1]{fontenc}    %
\usepackage[colorlinks,citecolor=black,urlcolor=black,linkcolor=black]{hyperref}       %
\usepackage{url}            %
\usepackage{booktabs}       %
\usepackage{amsfonts}       %
\usepackage{nicefrac}       %
\usepackage{microtype}      %
\usepackage[numbers,sort&compress]{natbib}
\usepackage{amsthm}
\usepackage{amsmath}
\usepackage[capitalize,nameinlink]{cleveref}
\usepackage{xcolor}
\usepackage{tabularx}
\usepackage{enumerate}
\usepackage{wrapfig}
\usepackage{subcaption}

\usepackage{tikz,pgfplots}
\pgfplotsset{compat=newest}
\usetikzlibrary{calc}
\usetikzlibrary{plotmarks,shapes,arrows,patterns,fadings}
\usetikzlibrary{shapes.misc}
\usepgfplotslibrary{fillbetween}
\newlength{\figurewidth}
\newlength{\figureheight}
\newlength{\blockwidth}

\tikzset{cross/.style={cross out, draw=black, minimum size=2*(#1-\pgflinewidth), inner sep=0pt, outer sep=0pt, line width=1pt}, cross/.default={3pt}}

\newcommand{\mbf}[1]{\mathbf{#1}}

\usepackage{algorithmic}
\AtBeginEnvironment{algorithmic}{}
\usepackage{algorithm}
\usepackage{bm}

\newcommand{\vzero}{\bm{0}}
\newcommand{\diff}{\,\mathrm{d}}

\newcommand{\vbeta}{\boldsymbol{\beta}}

\newcommand{\MSigma}{\mbf{\Sigma}}

\renewcommand{\mid}[0]{\,|\,}

\newcommand{\vw}{\mbf{w}}
\newcommand{\vx}{\mbf{x}}
\newcommand{\vy}{\mbf{y}}
\newcommand{\vz}{\mbf{z}}

\newcommand{\MH}{\mbf{H}}
\newcommand{\MI}{\mbf{I}}

\renewcommand{\paragraph}[1]{\smallskip\textbf{#1:}~}

\begin{document}
\title{Learning to Approximate Particle Smoothing Trajectories via Diffusion Generative Models}
\author{\IEEEauthorblockN{Ella Tamir}
\IEEEauthorblockA{Aalto University \\
Espoo, Finland\\
\parbox{.3\textwidth}{\centering ella.tamir@aalto.fi}}
\and
\IEEEauthorblockN{Arno Solin}
\IEEEauthorblockA{Aalto University\\
Espoo, Finland\\
\parbox{.3\textwidth}{\centering arno.solin@aalto.fi}}
}
\maketitle
\begin{abstract}
Learning dynamical systems from sparse observations is critical in numerous fields, including biology, finance, and physics. Even if tackling such problems is standard in general information fusion, it remains challenging for contemporary machine learning models, such as diffusion models. We introduce a method that integrates conditional particle filtering with ancestral sampling and diffusion models, enabling the generation of realistic trajectories that align with observed data. Our approach uses a smoother based on iterating a conditional particle filter with ancestral sampling to first generate plausible trajectories matching observed marginals, and learns the corresponding diffusion model. This approach provides both a generative method for high-quality, smoothed trajectories under complex constraints, and an efficient approximation of the particle smoothing distribution for classical tracking problems. We demonstrate the approach in time-series generation and interpolation tasks, including vehicle tracking and single-cell RNA sequencing data.
\end{abstract}

\IEEEpeerreviewmaketitle

\section{Introduction}
\label{sec:introduction}
Learning a time-series model based on sparse observations is a problem arising in various practical applications, in fields such as biology, finance, and physics. Past approaches using ordinary differential equations or stochastic differential equations with a neural network as the driving force or drift include for instance \cite{rubanova2019latent}, \cite{yildiz2019ode2vae} and \cite{li2020scalable}. Trajectory learning approaches for systems of sparse observations have been explored in \cite{lavenant2023mathematical}, while a Markov Chain Monte Carlo (MCMC) based particle smoothing approach was studied in \cite{wigren2019parameter}.

On the other hand, score-based diffusion models \cite{song2021scorebased} for generating complex data have demonstrated state-of-the-art performance for various data modalities, such as images, graphs, and video. Diffusion models pose a generative task as sampling from a simple distribution $\pi_0$ (often $\mathcal{N}(\vzero, \MI)$), and transforming the samples to an arbitrary data distribution through applying a learned Stochastic Differential Equation (SDE) model until time $T$---thus obtaining samples from a distribution $\pi_T$ which can be sampled, but not evaluated. The behaviour of the intermediate SDE system is often not of interest, but merely used as a vehicle for obtaining high-quality samples from $\pi_T$. For a setting where even the initial distribution $\pi_0$ is allowed to be a complex data distribution, iterative Schrödinger bridge methods such as the Iterative Proportional Fitting Procedure (IPFP)~\cite{debortoli2021diffusion} allow for learning the intermediate SDE system, with additional regularization through setting a reference process whose marginals the Schrödinger bridge should match as closely as possible, while adhering to the marginal constraints at $\pi_0$ and $\pi_T$.

We seek to combine the two problem settings, allowing for both sparse observations and complex constraints $\pi_0$ and $\pi_T$. To that end, we unite diffusion models with particle smoothing methods suited for complex data. The Conditional Particle Filter with Ancestor Sampling (CPF-AS, \cite{lindsten2014particle, svensson2015nonlinear}) combined with MCMC sampling provides an effective scheme for generating trajectories from the smoothing distribution, with theoretical guarantees of convergence. In \cite{verma2023variational}, an MCMC baseline was presented for parameter inference, as a comparison to more scalable variational inference based methods. The MCMC baseline applied CPF-AS in order to effectively evaluate the joint log likelihood over observations and parameters. Inspired by the baseline method, we adopt it to a modern deep learning setting, where in addition to performing inference under complex constraints (both sparse observations and terminal distributions), we learn a model which emulates the behaviour of the inference trajectories, while providing an SDE with an explicitly defined drift function to be used in Euler--Maruyama sampling (see, e.g., \cite{sarkka2019applied}).\looseness-1

\begin{figure}[t!]
  \centering\scriptsize
  \pgfplotsset{scale only axis}
  \setlength{\figureheight}{.5\columnwidth}  
  \newcommand{\tbox}[1]{\tikz\node[inner sep=0,draw=none]{#1};}
  \hfill
  \begin{subfigure}[b]{0.17\columnwidth}
    \centering
    \setlength{\figurewidth}{\textwidth}
    \tbox{
\begin{tikzpicture}

\definecolor{darkgray176}{RGB}{176,176,176}
\definecolor{steelblue31119180}{RGB}{31,119,180}

\begin{axis}[
height=\figureheight,
hide x axis,
hide y axis,
tick align=outside,
tick pos=left,
width=\figurewidth,
x grid style={darkgray176},
xmin=-1.29691396709952, xmax=0.0617578079571197,
xtick style={color=black},
y grid style={darkgray176},
ymin=-1.5, ymax=1.5,
ytick style={color=black}
]
\addplot [name path=f, semithick, steelblue31119180]
table {%
-5.65223898895306e-07 -1.5
-1.25048531309428e-06 -1.48492462311558
-2.66809134569364e-06 -1.46984924623116
-5.49023261502824e-06 -1.45477386934673
-1.08956765268561e-05 -1.43969849246231
-2.08544443562395e-05 -1.42462311557789
-3.84976747395986e-05 -1.40954773869347
-6.85450948914313e-05 -1.39447236180905
-0.000117718432716535 -1.37939698492462
-0.00019501445836329 -1.3643216080402
-0.000311660920428482 -1.34924623115578
-0.000480557309688903 -1.33417085427136
-0.000715040253125364 -1.31909547738693
-0.00102693432768393 -1.30402010050251
-0.00142405479808932 -1.28894472361809
-0.00190758393138089 -1.27386934673367
-0.00246997106186947 -1.25879396984925
-0.00309410678821488 -1.24371859296482
-0.00375440191149638 -1.2286432160804
-0.00442002678286707 -1.21356783919598
-0.00505999501214205 -1.19849246231156
-0.00564916369192233 -1.18341708542714
-0.00617378453056441 -1.16834170854271
-0.00663516910954897 -1.15326633165829
-0.00705041095411251 -1.13819095477387
-0.00744985950843982 -1.12311557788945
-0.0078719392787699 -1.10804020100503
-0.0083566556285373 -1.0929648241206
-0.0089394795708622 -1.07788944723618
-0.00964715680293979 -1.06281407035176
-0.0104964193162485 -1.04773869346734
-0.0114957989675306 -1.03266331658291
-0.0126499875661967 -1.01758793969849
-0.0139656255809268 -1.00251256281407
-0.0154570921652689 -0.987437185929648
-0.0171507973886631 -0.972361809045226
-0.0190866221103905 -0.957286432160804
-0.0213155356098818 -0.942211055276382
-0.0238931000917358 -0.92713567839196
-0.0268695498214301 -0.912060301507538
-0.0302782744767484 -0.896984924623116
-0.0341255295173883 -0.881909547738693
-0.0383846269868594 -0.866834170854271
-0.0429973795970851 -0.851758793969849
-0.0478840856238043 -0.836683417085427
-0.052961127910829 -0.821608040201005
-0.058162931080865 -0.806532663316583
-0.0634633450032842 -0.791457286432161
-0.0688911478183787 -0.776381909547739
-0.0745355565587321 -0.761306532663317
-0.0805401439887251 -0.746231155778895
-0.087086651487639 -0.731155778894472
-0.0943728666633267 -0.71608040201005
-0.102590092781551 -0.701005025125628
-0.11190529792382 -0.685929648241206
-0.122450965917813 -0.670854271356784
-0.134322781000954 -0.655778894472362
-0.147582720607534 -0.64070351758794
-0.162263956561894 -0.625628140703518
-0.178374650998501 -0.610552763819096
-0.195899916414736 -0.595477386934673
-0.214803779713923 -0.580402010050251
-0.235034554559847 -0.565326633165829
-0.25653656967649 -0.550251256281407
-0.279268644016109 -0.535175879396985
-0.303226052764951 -0.520100502512563
-0.328459697760664 -0.505025125628141
-0.355085382788158 -0.489949748743719
-0.383278156285404 -0.474874371859296
-0.413250915768922 -0.459798994974874
-0.44522107195276 -0.444723618090452
-0.479372035061567 -0.42964824120603
-0.515816436195287 -0.414572864321608
-0.554565636921815 -0.399497487437186
-0.595506841748481 -0.384422110552764
-0.638387063062547 -0.369346733668342
-0.682803588047969 -0.35427135678392
-0.728203195158128 -0.339195979899497
-0.773895393654526 -0.324120603015075
-0.819086047833026 -0.309045226130653
-0.862935223327089 -0.293969849246231
-0.904636952877001 -0.278894472361809
-0.943510690767675 -0.263819095477387
-0.979087439022327 -0.248743718592965
-1.01117068968624 -0.233668341708543
-1.03985499350148 -0.218592964824121
-1.06549290747662 -0.203517587939699
-1.0886124512688 -0.188442211055276
-1.10979921940895 -0.173366834170854
-1.12956697297513 -0.158291457286432
-1.14824532211283 -0.14321608040201
-1.16591132785988 -0.128140703517588
-1.18238305955395 -0.113065326633166
-1.19727853318891 -0.0979899497487438
-1.21012607724751 -0.0829145728643217
-1.22049666003302 -0.0678391959798996
-1.22812021187712 -0.0527638190954773
-1.23295037262083 -0.0376884422110553
-1.2351561591424 -0.0226130653266332
-1.23504141491656 -0.00753768844221114
-1.23291664485251 0.00753768844221092
-1.22896470210061 0.0226130653266332
-1.22314534950781 0.0376884422110553
-1.21517207571546 0.0527638190954773
-1.20457137172154 0.0678391959798994
-1.19080796238936 0.0829145728643215
-1.17343853737306 0.0979899497487438
-1.15224839553832 0.113065326633166
-1.12733234157343 0.128140703517588
-1.09909996416971 0.14321608040201
-1.06820898061983 0.158291457286432
-1.03545051593165 0.173366834170854
-1.00162082599474 0.188442211055276
-0.967412835305951 0.203517587939698
-0.93334992999205 0.21859296482412
-0.899768676143601 0.233668341708543
-0.866842288501209 0.248743718592965
-0.834627244364276 0.263819095477387
-0.803113343462474 0.278894472361809
-0.772261944843327 0.293969849246231
-0.742025323018134 0.309045226130653
-0.712348687025916 0.324120603015075
-0.683162578390376 0.339195979899497
-0.654375615621056 0.35427135678392
-0.625875882236016 0.369346733668342
-0.597544825249699 0.384422110552764
-0.569282094920168 0.399497487437186
-0.541035075296964 0.414572864321608
-0.512824263606347 0.42964824120603
-0.484755826033669 0.444723618090452
-0.457015526667378 0.459798994974874
-0.429843053905229 0.474874371859296
-0.40349128626895 0.489949748743719
-0.378179713763139 0.505025125628141
-0.354053632721233 0.520100502512563
-0.331159960326013 0.535175879396985
-0.309446545688842 0.550251256281407
-0.288785659097361 0.565326633165829
-0.269015669977182 0.580402010050251
-0.24998977976959 0.595477386934673
-0.231618649868079 0.610552763819095
-0.213895490442992 0.625628140703518
-0.196897115421152 0.640703517587939
-0.180761112611287 0.655778894472362
-0.165645667190624 0.670854271356784
-0.151682937137725 0.685929648241206
-0.138938137017748 0.701005025125628
-0.127384513454949 0.71608040201005
-0.116899946385754 0.731155778894472
-0.107285322015792 0.746231155778895
-0.0982996007694292 0.761306532663316
-0.089702914672879 0.776381909547739
-0.0812978107237216 0.791457286432161
-0.072959984574451 0.806532663316583
-0.0646529798531235 0.821608040201005
-0.0564253998558438 0.836683417085427
-0.0483930963578315 0.851758793969849
-0.0407116402508412 0.866834170854271
-0.0335456116579112 0.881909547738693
-0.0270408314490267 0.896984924623116
-0.0213039777570885 0.912060301507537
-0.0163917277753212 0.92713567839196
-0.0123093042080258 0.942211055276382
-0.00901659540933088 0.957286432160804
-0.00643910588446524 0.972361809045226
-0.00448086600392074 0.987437185929648
-0.00303688870519501 1.00251256281407
-0.00200353204334078 1.01758793969849
-0.00128595339954417 1.03266331658291
-0.000802543102461981 1.04773869346734
-0.000486712320367301 1.06281407035176
-0.000286669268517838 1.07788944723618
-0.000163885283526804 1.0929648241206
-9.08864318699459e-05 1.10804020100502
-4.8867176058817e-05 1.12311557788945
-2.54603749978153e-05 1.13819095477387
-1.28477076794567e-05 1.15326633165829
-6.27623035003515e-06 1.16834170854271
-2.96687687985614e-06 1.18341708542714
-1.35662196328189e-06 1.19849246231156
-5.99824302006473e-07 1.21356783919598
-2.56364435703279e-07 1.2286432160804
-1.05885664486952e-07 1.24371859296482
-4.2252551056874e-08 1.25879396984925
-1.62857683175473e-08 1.27386934673367
-6.06202360004464e-09 1.28894472361809
-2.17874187851212e-09 1.30402010050251
-7.55973592741676e-10 1.31909547738693
-2.53199132290158e-10 1.33417085427136
-8.18505939382549e-11 1.34924623115578
-2.55353118413189e-11 1.3643216080402
-7.68743569206128e-12 1.37939698492462
-2.23310448194339e-12 1.39447236180905
-6.25884708465051e-13 1.40954773869347
-1.6924338116983e-13 1.42462311557789
-4.41508119453071e-14 1.43969849246231
-1.11110681470656e-14 1.45477386934673
-2.69740406972033e-15 1.46984924623116
-6.31676077344965e-16 1.48492462311558
-1.42688218027483e-16 1.5
};

\path[name path=axis] (axis cs:0,-1.5) -- (axis cs:0,1.5);    
        
\addplot[thick,color=steelblue31119180,fill=steelblue31119180,fill opacity=0.1]
    fill between[
        of=f and axis,
    ];        

\node[align=center] at (axis cs:-.5,0) {$\pi_0$};

\end{axis}

\end{tikzpicture}}
  \end{subfigure}
  \begin{subfigure}[b]{0.6\columnwidth}
    \centering
    \setlength{\figurewidth}{\textwidth}
    \tbox{\input{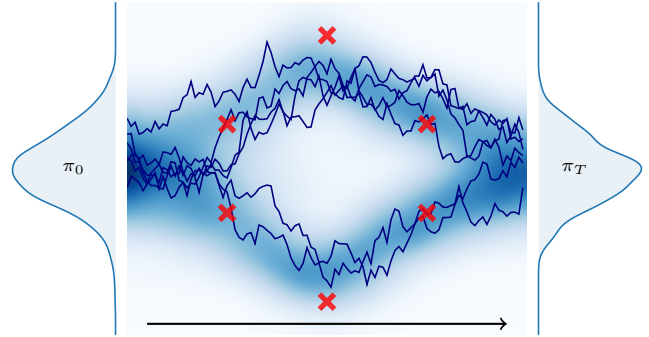}}
  \end{subfigure}
  \begin{subfigure}[b]{0.17\columnwidth}
    \centering
    \setlength{\figurewidth}{\textwidth}  
    \tbox{
\begin{tikzpicture}

\definecolor{darkgray176}{RGB}{176,176,176}
\definecolor{steelblue31119180}{RGB}{31,119,180}

\begin{axis}[
height=\figureheight,
hide x axis,
hide y axis,
tick align=outside,
tick pos=left,
width=\figurewidth,
x grid style={darkgray176},
xmin=-0.0702848634286122, xmax=1.47598213200086,
xtick style={color=black},
y grid style={darkgray176},
ymin=-1.5, ymax=1.5,
ytick style={color=black}
]
\addplot [name path=f, semithick, steelblue31119180]
table {%
5.27842318491459e-25 -1.5
4.36683961842256e-24 -1.48492462311558
3.45388040190472e-23 -1.46984924623116
2.6117084978044e-22 -1.45477386934673
1.88807779409699e-21 -1.43969849246231
1.30494849430986e-20 -1.42462311557789
8.62275056939648e-20 -1.40954773869347
5.44725962434634e-19 -1.39447236180905
3.28996418162373e-18 -1.37939698492462
1.8997061160109e-17 -1.3643216080402
1.0487343167685e-16 -1.34924623115578
5.53516309402752e-16 -1.33417085427136
2.79308314638912e-15 -1.31909547738693
1.34750101646176e-14 -1.30402010050251
6.21541724060565e-14 -1.28894472361809
2.74102291360315e-13 -1.27386934673367
1.1557472087852e-12 -1.25879396984925
4.65939158496785e-12 -1.24371859296482
1.79606607490051e-11 -1.2286432160804
6.61996237799093e-11 -1.21356783919598
2.33316670740312e-10 -1.19849246231156
7.86347465383178e-10 -1.18341708542714
2.53447980978864e-09 -1.16834170854271
7.81273714203566e-09 -1.15326633165829
2.30356901291806e-08 -1.13819095477387
6.49737802095119e-08 -1.12311557788945
1.7534117628063e-07 -1.10804020100503
4.5282159374945e-07 -1.0929648241206
1.11938552976523e-06 -1.07788944723618
2.64961475608537e-06 -1.06281407035176
6.0077761901049e-06 -1.04773869346734
1.30555451627447e-05 -1.03266331658291
2.72084273778618e-05 -1.01758793969849
5.44228799814109e-05 -1.00251256281407
0.000104580060605617 -0.987437185929648
0.000193291834315881 -0.972361809045226
0.000344096893281354 -0.957286432160804
0.00059095989839925 -0.942211055276382
0.000980960929183715 -0.92713567839196
0.00157707904934292 -0.912060301507538
0.00246100694181991 -0.896984924623116
0.00373590979004712 -0.881909547738693
0.00552886702376999 -0.866834170854271
0.0079923654788254 -0.851758793969849
0.0113037298046695 -0.836683417085427
0.015661014108676 -0.821608040201005
0.0212739354848669 -0.806532663316583
0.0283490863823735 -0.791457286432161
0.0370698191446476 -0.776381909547739
0.0475724728644057 -0.761306532663317
0.0599215766669041 -0.746231155778895
0.0740871522435265 -0.731155778894472
0.0899275085927974 -0.71608040201005
0.107181407817372 -0.701005025125628
0.1254742475204 -0.685929648241206
0.144343217705559 -0.670854271356784
0.163284884964033 -0.655778894472362
0.181824239385363 -0.64070351758794
0.199597223188841 -0.625628140703518
0.216431394847054 -0.610552763819096
0.232405158537544 -0.595477386934673
0.247867876437705 -0.580402010050251
0.263411775224702 -0.565326633165829
0.279799305620395 -0.550251256281407
0.297861733873992 -0.535175879396985
0.318391438238641 -0.520100502512563
0.342049171582993 -0.505025125628141
0.369299492372865 -0.489949748743719
0.400376739348933 -0.474874371859296
0.435275436529352 -0.459798994974874
0.473756497624672 -0.444723618090452
0.515364392284594 -0.42964824120603
0.559457345553174 -0.414572864321608
0.605257492073522 -0.399497487437186
0.651926282567282 -0.384422110552764
0.698661325700269 -0.369346733668342
0.74479774771943 -0.35427135678392
0.7898868594736 -0.339195979899497
0.833724268540946 -0.324120603015075
0.876311671874611 -0.309045226130653
0.917758525925046 -0.293969849246231
0.958153360702425 -0.278894472361809
0.997449501280314 -0.263819095477387
1.03540893633353 -0.248743718592965
1.07162988265615 -0.233668341708543
1.1056541575227 -0.218592964824121
1.13712046248237 -0.203517587939699
1.16591032331869 -0.188442211055276
1.19223222570519 -0.173366834170854
1.21660749485407 -0.158291457286432
1.23975291136137 -0.14321608040201
1.26238924326129 -0.128140703517588
1.28502983393629 -0.113065326633166
1.30781012478576 -0.0979899497487438
1.33040497857641 -0.0829145728643217
1.35205111953016 -0.0678391959798996
1.37165816237832 -0.0527638190954773
1.38796648467675 -0.0376884422110553
1.39970288603861 -0.0226130653266332
1.40569726857224 -0.00753768844221114
1.40494886753749 0.00753768844221092
1.39665694018013 0.0226130653266332
1.38024636568285 0.0376884422110553
1.35541663973693 0.0527638190954773
1.32222459699385 0.0678391959798994
1.28118537130325 0.0829145728643215
1.23335438159827 0.0979899497487438
1.18034545983424 0.113065326633166
1.12425071084831 0.128140703517588
1.06745316468763 0.14321608040201
1.01235493427089 0.158291457286432
0.961070564622238 0.173366834170854
0.915148708558146 0.188442211055276
0.875381269096012 0.203517587939698
0.841739542936353 0.21859296482412
0.813447989477099 0.233668341708543
0.789176254003658 0.248743718592965
0.767306438623361 0.263819095477387
0.746220178333035 0.278894472361809
0.724550438986743 0.293969849246231
0.701354786061356 0.309045226130653
0.67618670103951 0.324120603015075
0.649064563868212 0.339195979899497
0.620358979711337 0.35427135678392
0.590633631161193 0.369346733668342
0.560479934773372 0.384422110552764
0.530381163243993 0.399497487437186
0.500629655529839 0.414572864321608
0.471305321445854 0.42964824120603
0.442309192976897 0.444723618090452
0.413435503394476 0.459798994974874
0.384461033180258 0.474874371859296
0.355230826281212 0.489949748743719
0.32572339318458 0.505025125628141
0.296084544303574 0.520100502512563
0.266625765193118 0.535175879396985
0.237789689649138 0.550251256281407
0.210091012522407 0.565326633165829
0.184045260107969 0.580402010050251
0.160099325766367 0.595477386934673
0.138576115958431 0.610552763819095
0.119641445556711 0.625628140703518
0.103295821813332 0.640703517587939
0.0893887429647624 0.655778894472362
0.0776499729260323 0.670854271356784
0.0677312506468634 0.685929648241206
0.059252328372027 0.701005025125628
0.0518459967436873 0.71608040201005
0.0451971654913282 0.731155778894472
0.039071288181126 0.746231155778895
0.0333282018981331 0.761306532663316
0.0279194515744788 0.776381909547739
0.0228703201587753 0.791457286432161
0.0182511933622132 0.806532663316583
0.0141452682780775 0.821608040201005
0.0106199975029785 0.836683417085427
0.00770789777833222 0.851758793969849
0.00539917146639644 0.866834170854271
0.00364519934920034 0.881909547738693
0.00236948509761398 0.896984924623116
0.00148165927976231 0.912060301507537
0.000890626190929904 0.92713567839196
0.000514326677303419 0.942211055276382
0.000285209300119042 0.957286432160804
0.000151804910732888 0.972361809045226
7.75259402336999e-05 0.987437185929648
3.79759747900862e-05 1.00251256281407
1.78381259398936e-05 1.01758793969849
8.03265132334926e-06 1.03266331658291
3.46688971200966e-06 1.04773869346734
1.43385068841963e-06 1.06281407035176
5.68159186567241e-07 1.07788944723618
2.15657593500948e-07 1.0929648241206
7.84008367614959e-08 1.10804020100502
2.7294553168525e-08 1.12311557788945
9.09858224609749e-09 1.13819095477387
2.90376529721886e-09 1.15326633165829
8.87137746823138e-10 1.16834170854271
2.59428594767113e-10 1.18341708542714
7.26107286832257e-11 1.19849246231156
1.94492126600893e-11 1.21356783919598
4.98524580610438e-12 1.2286432160804
1.22270644782609e-12 1.24371859296482
2.86932112043768e-13 1.25879396984925
6.44213764595745e-14 1.27386934673367
1.38372389001852e-14 1.28894472361809
2.84324412302826e-15 1.30402010050251
5.5885949812234e-16 1.31909547738693
1.0507395719818e-16 1.33417085427136
1.88961468823381e-17 1.34924623115578
3.25027665324099e-18 1.3643216080402
5.34713099724432e-19 1.37939698492462
8.41318595959636e-20 1.39447236180905
1.26597812037381e-20 1.40954773869347
1.82182174258227e-21 1.42462311557789
2.50719675321065e-22 1.43969849246231
3.2996173674932e-23 1.45477386934673
4.15261783468837e-24 1.46984924623116
4.99752429620114e-25 1.48492462311558
5.7511470047687e-26 1.5
};

\path[name path=axis] (axis cs:0,-1.5) -- (axis cs:0,1.5);    
        
\addplot[thick,color=steelblue31119180,fill=steelblue31119180,fill opacity=0.1]
    fill between[
        of=f and axis,
    ];        

\node[align=center] at (axis cs:.5,0) {$\pi_T$};

\end{axis}

\end{tikzpicture}}
  \end{subfigure}
  \hfill
  \caption{Trajectories from an SDE with the neural network drift model learned from smoothing trajectories over a zero drift model with the constraints $\pi_0 = \pi_T = \mathcal{N}(0, 1)$, with observations \protect\tikz[baseline=-.5ex]\protect\node[cross=5.5pt,line width=2.5pt,red]{}; forcing the trajectories to split into two modes.}
   \label{fig:teaser}
   \vspace*{-2em}
\end{figure}

Recent work by \cite{tamir2023transport} combines a differentiable particle filter with iterative bridge methods, in order to account for both the terminal constraints and sparse observations from the latent marginals. We position our method as a non-iterative version of \cite{tamir2023transport}, where the MCMC particle smoother allows us to incorporate data without the need for backwards trajectory simulation. The key benefit of our approach lies in enabling faster sampling from the particle smoothing distribution after the MCMC chain for CPF-AS has converged via using the learned diffusion model, while providing a neural SDE model.

Our contributions are summarized as follows.
\begin{enumerate}[(i)]
    \item \textbf{Inference:} We propose a tailored approach combining Conditional Particle Filtering with Ancestral Sampling to account for multiple observations and terminal constraints.
    \item \textbf{Learning:} We apply learning strategies used for iterative Schrödinger Bridge methods to learn a model matching smoother trajectories, providing a scalable neural approximation of an MCMC particle smoother.
    \item \textbf{Experiments:} We provide strong experimental results for simulated and empirical data including vehicle tracking and single-cell RNA data, both in time-series generation and interpolation tasks.
\end{enumerate}

\section{Related work}
\label{sec:related}
\paragraph{Conditional Particle Filtering with Ancestral Sampling}
The CPF-AS smoother was formulated originally in \cite{svensson2015nonlinear}, where it was presented for the purpose of accurate particle smoothing together with derivations of convergence results for the smoother. 
Obtaining good performance in the Conditional Particle Filter method for trajectory generation is a non-trivial task. It has been extensively studied in recent work \cite{karppinen2021cpf, karppinen2023cpf} to extend the original CPF-AS to various settings such as cases where the initial latent state distribution is a diffusion instead of a single point, or in using bridge backward sampling instead of ancestral sampling. The approach in \cite{chopin2022resampling} describes resampling schemes for weak observations and explores dense time slices. In recent applications of CPF-AS, \cite{kok2023raoblackwellized} uses modification of CPF-AS with Rao--Blackwellized particle dynamics for a SLAM problem, while \cite{wigren2019parameter} explores particle Gibbs with Ancestral Sampling, a conditional Sequential Monte Carlo (SMC) approach similar to CPF-AS for parameter estimation. These sequantial Monte Carlo approaches provide an alternative to variational inference and non-linear filtering (see \cite{solin2021scalable}) approaches often favoured in real-time approaches.
  
\paragraph{Diffusion Models and Schrödinger Bridges}
Diffusion models \cite{ho2020denoising, song2021scorebased} have been successfully applied to a multitude of generative problem settings. Conditioning diffusion models based on class label was proposed already in \cite{song2021scorebased}, but more complex data conditioning has been studied for instance in \cite{watson2023novel}, where 3D view synthesis is conditioned on 2D images. Other conditional settings such as \cite{cardoso2023monte} and \cite{trippe2023diffusion} consider graph generation for molecules, where the motif-scaffolding problem is solved through using an SMC algorithm for conditional generation and guidance of the diffusion.

For an extension of diffusion model approaches to cases where both the initial and terminal distribution are unknown, iterative approaches such as IPFP for solving Schrödinger bridges were proposed in \cite{debortoli2021diffusion} and \cite{vargas2021solving}. IPFP is based on solving half-bridge problems, and has been applied to optimal transport problems for images and biological data.

\paragraph{Diffusion Models, Control, and Data Assimilation}
The concept of applying ideas from stochastic control to diffusion model has been studied for instance in \cite{berner2022optimal,chen2023generative,cardoso2023monte}, while in \cite{maoutsa2022deterministic} a deterministic particle filter is used to generate samples from nonlinear dynamical systems under observations. In \cite{tamir2023transport}, a particle filter with differential resampling is applied to a constrained generative task, combining the iterative half-bridge approach from \cite{debortoli2021diffusion} with particle filter based data assimilation from \cite{maoutsa2022deterministic}. Further connections between Schrödinger bridges and optimal control have been explored in \cite{chen2022likelihood} using the forward-backward SDE formulation of Schrödinger bridges, and in \cite{caluya2020reflected} and \cite{caluya2019wasserstein} for additional constraints on the controlled system with nonlinear drift. Incorporating information from observations to a known dynamical systems has been explored in the field of data assimilation \citep{asch2016data}, for instance combining Schrödinger bridges with data assimilation in \cite{reich2019data}.

\section{Methods}
\label{sec:methods}
Our proposed method learns to approximate the particle smoothing distribution via utilizing samples generated through MCMC over Conditional Particle Filtering with Ancestral Sampling (CPF-AS, \cite{svensson2015nonlinear}). The learning step consists of applying trajectory reversion techniques from the diffusion model and Schrödinger Bridge literature. 
First, we present our particle smoothing notation in \cref{sec:smoothing}, and briefly introduce diffusion models in \cref{sec:diffusion}, and to MCMC smoothing using CPF-AS in \cref{sec:cpfas}. In \cref{sec:algorithms}, we explain how we condition the CPF-AS on multiple observations and a terminal constraint to adapt it to our setting, and in \cref{sec:learning} combine trajectory learning with smoothing trajectories.  \looseness-1

\subsection{Particle Smoothing and Notation}\label{sec:smoothing}
We consider the state-space model
\begin{align*}
& \diff \vx_t = f(\vx_t, t) \diff t + g(t)^2 \diff \vbeta_t,  \\
& \vx_0 \sim \pi_{0}, \\
& \vy_{t_k} \sim \mathcal{N}(\MH \vx_{t_k}, \MSigma), \quad \forall k\in \{1,\ldots,K \},
\end{align*}
where $t \in [0, T]$ is the time horizon, $\pi_0$ is a bounded initial distribution, $\{ t_k\}_{k=1}^K$ are observation times, and for each $k$, $\vy_{t_k} \in \mathbb{R}^{d_y}$ and $\forall t \in [0, T]$, $\vx_t \in \mathbb{R}^{d_x}$, and $\MH \in \mathbb{R}^{d_x \times d_y}$. In other words, we have restricted to SDEs with nonlinear dynamics and linear Gaussian observation models, and are interested in the continuous-discrete setting. In practice, we will resort to Euler--Maruyama simulations of the SDE. 

We find the conditional state distribution $p(\vx_{t}  \mid \{\vy_{t_k}\}_{k=1}^K)$ through simulating particle trajectories through a particle filter. We denote by $w_{t}^j$ the particle weight of the $j\textsuperscript{th}$ particle at time $t$ and by $\vx_{t}^j$ its corresponding particle, and generate $N$ particles.

\begin{figure}[t!]
  \centering\small
  \pgfplotsset{scale only axis}
  \setlength{\figurewidth}{0.45\textwidth}
  \setlength{\figureheight}{0.4\figurewidth}  
  \begin{subfigure}[b]{\figurewidth}
    \centering
\begin{tikzpicture}

\definecolor{darkgray176}{RGB}{176,176,176}

\begin{axis}[
height=\figureheight,
hide x axis,
hide y axis,
tick align=outside,
tick pos=left,
width=\figurewidth,
x grid style={darkgray176},
xmin=0, xmax=1,
xtick style={color=black},
y grid style={darkgray176},
ymin=-1.5, ymax=1.5,
ytick style={color=black}
]
\addplot [semithick, blue, dashed]
table {%
0 0.399354604865631
0.00999999977648258 0.300490354980549
0.0199999995529652 0.21634074100407
0.0299999993294477 0.214935735777227
0.0399999991059303 0.170448825672476
0.0499999970197678 0.0732947597058696
0.0599999986588955 0.259359509662001
0.0700000002980232 0.135674909666388
0.0799999982118607 0.146781167924611
0.0899999961256981 0.0296682258846445
0.0999999940395355 0.0867149932088537
0.109999999403954 0.096135599516539
0.119999997317791 0.13068738018742
0.129999995231628 0.152060227446465
0.140000000596046 0.112511286489872
0.149999991059303 -0.0636551734281497
0.159999996423721 -0.261827891714664
0.170000001788139 -0.328995389885994
0.179999992251396 -0.247193573064895
0.189999997615814 -0.202363705284687
0.199999988079071 -0.236130354948135
0.209999993443489 -0.186602482594343
0.219999998807907 -0.226207850523086
0.229999989271164 -0.374219609446617
0.239999994635582 -0.472898548491092
0.25 -0.699783450014682
0.259999990463257 -0.555686121828647
0.269999980926514 -0.411886503942104
0.280000001192093 -0.333359247095676
0.28999999165535 -0.313915696017237
0.299999982118607 -0.332283526681157
0.310000002384186 -0.49837945403883
0.319999992847443 -0.455059852473231
0.329999983310699 -0.635943289868327
0.340000003576279 -0.788065504304858
0.349999994039536 -0.853922557107897
0.359999984502792 -0.983692389003726
0.370000004768372 -0.916395414642783
0.379999995231628 -1.11358356850574
0.389999985694885 -1.10988288700932
0.399999976158142 -1.07521209821457
0.409999996423721 -1.20326637432331
0.419999986886978 -1.19402336191738
0.429999977350235 -1.34032266449535
0.439999997615814 -1.33651364107968
0.449999988079071 -1.23283383508565
0.459999978542328 -1.15522568186166
0.469999998807907 -1.11815310527922
0.479999989271164 -1.24262650569321
0.489999979734421 -1.18697123115303
0.5 -1.13882835467698
0.509999990463257 -1.20836992820145
0.519999980926514 -1.17206358050348
0.530000030994415 -1.03629308079721
0.540000021457672 -0.961756161203401
0.550000011920929 -0.990972899621146
0.560000061988831 -0.93086941984834
0.570000052452087 -0.859635499095337
0.580000042915344 -0.928426128601449
0.590000033378601 -0.838042226409333
0.600000023841858 -0.847909734373705
0.610000014305115 -0.776154943531172
0.620000004768372 -0.85465236825289
0.629999995231628 -0.809753940230029
0.639999985694885 -0.909424782519953
0.649999976158142 -0.717311874752657
0.660000026226044 -0.566170559054034
0.670000016689301 -0.620864153628962
0.680000007152557 -0.709158220117705
0.690000057220459 -0.700632756365256
0.700000047683716 -0.561974545730071
0.710000038146973 -0.60758751795979
0.720000028610229 -0.477237866980748
0.730000019073486 -0.430139360738711
0.740000009536743 -0.315491130364852
0.75 -0.20895243228765
0.759999990463257 -0.0447134131321478
0.769999980926514 0.0351900702586603
0.780000030994415 0.218026096211953
0.790000021457672 0.200938126282973
0.800000011920929 0.268539374904914
0.810000002384186 0.288107147651
0.819999992847443 0.244005663531108
0.830000042915344 0.236419881904705
0.840000033378601 0.337067100430114
0.850000023841858 0.231908279681785
0.860000014305115 0.305339585269077
0.870000004768372 0.435256938660724
0.879999995231628 0.40839387218116
0.889999985694885 0.326849987009747
0.900000035762787 0.487835888961537
0.910000026226044 0.497478292534335
0.920000016689301 0.401199669787867
0.930000007152557 0.400585031522179
0.939999997615814 0.310325941456223
0.949999988079071 0.397868199659253
0.960000038146973 0.276916472626591
0.970000028610229 0.205679042650128
0.980000019073486 0.218253486243749
0.990000009536743 0.35142608770516
};
\addplot [draw=red, fill=red!50, mark=x, only marks, mark options={solid}, mark size=4pt, opacity=0.8, line width=2pt]
table {%
0.25 -0.4
0.25 0.4
0.5 -1.2
0.5 1.2
0.75 0.4
0.75 -0.4
};

\end{axis}

\end{tikzpicture}
    \captionsetup{justification=centering}
    \caption{$10\textsuperscript{th}$ reference trajectory}
  \end{subfigure}\\[1em]
   \begin{subfigure}[b]{\figurewidth}
    \centering
    \input{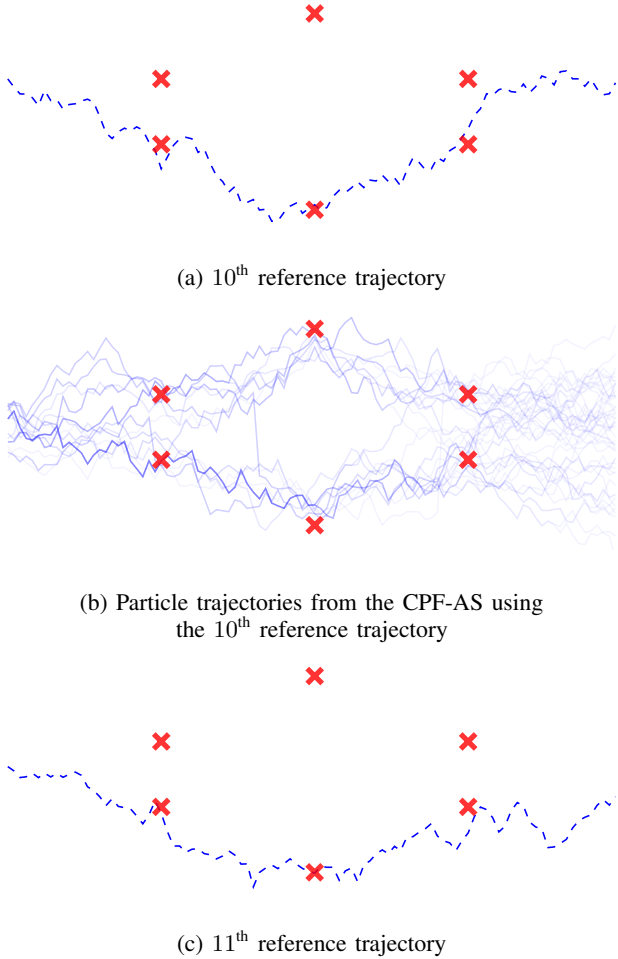}
    \captionsetup{justification=centering}
    \caption{Particle trajectories from the CPF-AS using \\ the $10\textsuperscript{th}$ reference trajectory}
  \end{subfigure}\\[1em]
  \begin{subfigure}[b]{\figurewidth}
    \centering
\begin{tikzpicture}

\definecolor{darkgray176}{RGB}{176,176,176}

\begin{axis}[
height=\figureheight,
hide x axis,
hide y axis,
tick align=outside,
tick pos=left,
width=\figurewidth,
x grid style={darkgray176},
xmin=0, xmax=1,
xtick style={color=black},
y grid style={darkgray176},
ymin=-1.5, ymax=1.5,
ytick style={color=black}
]
\addplot [semithick, blue, dashed]
table {%
0 0.0929637728948768
0.00999999977648258 0.0398358321507152
0.0199999995529652 -0.0414729424993817
0.0299999993294477 -0.031826002454323
0.0399999991059303 0.0126110297386225
0.0499999970197678 -0.0427774357850973
0.0599999986588955 0.013289311969715
0.0700000002980232 -0.0337969045396272
0.0799999982118607 0.0104044173602637
0.0899999961256981 -0.038517502998871
0.0999999940395355 -0.00120983452157417
0.109999999403954 0.0252491217796381
0.119999997317791 -0.0145978684778635
0.129999995231628 -0.154664730584187
0.140000000596046 -0.200262991642517
0.149999991059303 -0.276921322500271
0.159999996423721 -0.383772140299839
0.170000001788139 -0.301508901571793
0.179999992251396 -0.393924033200306
0.189999997615814 -0.409452189622385
0.199999988079071 -0.424158208934528
0.209999993443489 -0.510706149903995
0.219999998807907 -0.6073516988437
0.229999989271164 -0.498369009165031
0.239999994635582 -0.32676822033332
0.25 -0.479773790506584
0.259999990463257 -0.700345889596206
0.269999980926514 -0.879274458555443
0.280000001192093 -0.878222342511577
0.28999999165535 -0.956895392319125
0.299999982118607 -0.917738690903348
0.310000002384186 -0.912250326877398
0.319999992847443 -0.971524456446928
0.329999983310699 -0.985581214470727
0.340000003576279 -1.03631902758585
0.349999994039536 -1.00039921973453
0.359999984502792 -0.938395253313643
0.370000004768372 -0.989741868270499
0.379999995231628 -1.09994339097844
0.389999985694885 -1.09645788318099
0.399999976158142 -1.37846027261199
0.409999996423721 -1.20128191954078
0.419999986886978 -1.14848421088041
0.429999977350235 -1.11046019515575
0.439999997615814 -1.15498004055203
0.449999988079071 -1.09440918213309
0.459999978542328 -1.03546555659355
0.469999998807907 -1.09898732921661
0.479999989271164 -1.15236966258467
0.489999979734421 -1.18621627187909
0.5 -1.21017617120029
0.509999990463257 -1.12496657802345
0.519999980926514 -1.13900407443376
0.530000030994415 -1.13779310735495
0.540000021457672 -1.29824413212569
0.550000011920929 -1.18199626954825
0.560000061988831 -1.14718061911495
0.570000052452087 -1.31654633807571
0.580000042915344 -1.10833931195171
0.590000033378601 -1.06072443577082
0.600000023841858 -0.97516410651238
0.610000014305115 -0.925982555771184
0.620000004768372 -0.973002640927148
0.629999995231628 -0.882238744223428
0.639999985694885 -0.935859049135757
0.649999976158142 -0.770476485067916
0.660000026226044 -0.764190675156129
0.670000016689301 -0.709774681763661
0.680000007152557 -0.66817302389623
0.690000057220459 -0.603738733963979
0.700000047683716 -0.75753086193086
0.710000038146973 -0.815233231918824
0.720000028610229 -0.840375085757983
0.730000019073486 -0.934981150017989
0.740000009536743 -0.812516821013701
0.75 -0.718741578446639
0.759999990463257 -0.52126152573706
0.769999980926514 -0.402234075056327
0.780000030994415 -0.433708222048295
0.790000021457672 -0.533112951951039
0.800000011920929 -0.706277394132627
0.810000002384186 -0.636871257262242
0.819999992847443 -0.447278255062116
0.830000042915344 -0.34728887064935
0.840000033378601 -0.386826397316468
0.850000023841858 -0.584569172399056
0.860000014305115 -0.646053814725888
0.870000004768372 -0.666949670227182
0.879999995231628 -0.897458847077501
0.889999985694885 -0.890165772685719
0.900000035762787 -0.847078837851
0.910000026226044 -0.797311868915272
0.920000016689301 -0.659205612191868
0.930000007152557 -0.569390278706265
0.939999997615814 -0.527508531162453
0.949999988079071 -0.497437894741011
0.960000038146973 -0.381315783062888
0.970000028610229 -0.3118144948277
0.980000019073486 -0.37813399026723
0.990000009536743 -0.275720879832221
};
\addplot [draw=red, fill=red!50, mark=x, only marks, mark options={solid}, mark size=4pt, opacity=0.8, line width=2pt]
table {%
0.25 -0.4
0.25 0.4
0.5 -1.2
0.5 1.2
0.75 0.4
0.75 -0.4
};
\end{axis}

\end{tikzpicture}
    \captionsetup{justification=centering}
    \caption{$11\textsuperscript{th}$ reference trajectory}
  \end{subfigure}
  \caption{In an iteration of the CPF-AS smoother, the reference trajectory (top) is used to create conditional particle filtering trajectories (middle), from which a new reference trajectory is sampled based on weights on the last time step (bottom). The final weights are computed  based on a Kernel Density Estimate (KDE) over the distribution $\pi_T = \mathcal{N}(0, 1)$. The trajectories are sampled based on intermediate observations and the reference trajectory, resulting in multiple trajectories in the middle plot following roughly the previous reference.}
  \label{fig:cpfas}
  \vspace*{-1em}
\end{figure}

\subsection{Diffusion Models and Schrödinger Bridges}
\label{sec:diffusion}
Given two distributions, $\pi_0$ and $\pi_T$  we seek to find the drift functions $f_{\theta}, b_{\phi}: \mathbb{R}^d \times [0, T] \to \mathbb{R}^d$ such that the measure $\mathbb{P}$ defined by the marginals of the forward diffusion
\begin{equation}
\diff \vx_t = f_{\theta}(\vx_t, t)\diff t + g(t) \diff \vbeta_t \quad \vx_0 \sim \pi_0,
\end{equation}
satisfies $\mathbb{P}_T= \pi_T$, and the backward diffusion
\begin{equation}\label{eq:bwdsde}
\diff \vx_t = b_{\phi}(\vx_t, t)\diff t + g(t) \diff \vbeta_t \quad \vx_0 \sim \pi_T
\end{equation}
satisfies $\mathbb{P}_0 = \pi_0$. Denote the density function of the marginal distribution of $\mathbb{P}$ at time $t$ by $p_t(\vx)$. In diffusion models, the initial distribution $\pi_T$ is set to $\mathcal{N}(0, I)$ and the model $b_{\phi}$ is fixed to a linear function, noising the data distribution $\pi_T$. The reverse diffusion $f_{\theta}$ from noise to data is equal to
\begin{equation}
f_{\theta}(\vx_t, t) = b_{\phi}(\vx_t, t) - g(t)^2\nabla \ln p_t(\vx_t, t),
\end{equation}
that is, we can learn to generate data from noise (or $\pi_0$ from $\pi_T$) through reversing the dynamics via learning to approximate the score function $\nabla \ln p_t(\vx_t, t)$ based on the trajectories. 
More generally, for Schrödinger bridges \citep{schrod1932surla, leonard2014schrodi},  we define a reference measure $\mathbb{Q}$ on $\mathbb{R}^d \times [0, T]$,
and seek to minimize the Kullback--Leibler divergence $\mathrm{KL}(\mathbb{P}, \mathbb{Q})$. The reference measure may be used to encode prior information, for instance it may be defined through simulation of an SDE with drift $f_0$, as in \cite{debortoli2021diffusion}. In order to solve the Kullback--Leibler minimization problem while adhering to the terminal constraints of matching the distributions $\pi_0$ and $\pi_T$, \cite{debortoli2021diffusion} applies an iterative half-bridge method \cite{ruschendorf1995convergence}, alternating between the minimization targets
\begin{equation}\label{eq:half-bridge}
\min_{\mathbb{P}_{2i} \in \mathbb{P}(\cdot, \pi_T)}\hspace*{-1em}\mathrm{KL}(\mathbb{P}_{2i} \mid \mathbb{P}_{2i-1}) \quad \text{ and } \hspace*{-1em} \min_{\mathbb{P}_{2i+1} \in \mathbb{P}(\pi_0, \cdot)}\hspace*{-1em}\mathrm{KL}(\mathbb{P}_{2i+1} \mid \mathbb{P}_{2i}),
\end{equation}
where we denote by $\mathbb{P}(\cdot, \pi_T)$ measures which match $\pi_T$ at time $T$, and by $\mathbb{P}(\pi_0, \cdot)$ measures which match $\pi_0$ at time $0$.
For each of the minimization targets in \cref{eq:half-bridge}, the learning step consists of reversing trajectory dynamics via a mean matching objective. Suppose that the backwards drift $b_{\phi}$ is fixed, and we learn the forward drift $f_{\theta}$. Then the loss function is set to
\begin{multline}
\mathcal{L}(\phi) = \sum_{i=1}^N \sum_{j=1}^{N_T} \|b_{\phi}
(\vx_t, t)\Delta_j- (\vx_{t_{j+1}}^i + f_{\theta}(\vx_{t_{j+1}}^i, t_j) \Delta_j\\
- \vx_{t_j}^i - f_{\theta}(\vx_{t_{j}}^i, t_j)\Delta_j)\|^2,
\end{multline}
where $\{ \vx_{t}^i\}_{i=1}^N$ are the simulated particles from the SDE in \cref{eq:bwdsde} and $\Delta_j$ denotes the time step length at $t_j$, and $N_T$ is the number of steps in the discretization.
In other words, the forward drift should match the change in mean set by the backward model. Instead of directly comparing subsequent particles $\vx_{t_j}^i$ to $\vx_{t_{j-1}}^i$, the loss function considers the Gaussian distribution $\mathcal{N}(\vx_{t_{j-1}}^i + f_{\theta}(\vx_{t_{j-1}}^i, t_{j-1}), \Delta_{k}g(t_{j})^2 \MI)$ from which $\vx_{t_j}^i$ is sampled.

\begin{algorithm}[tb]
   \caption{Generating the first reference trajectory}
   \label{alg:init}
   \renewcommand{\algorithmiccomment}[1]{\hfill\textcolor{gray}{\(\triangleright\) #1}}
\begin{algorithmic}
   \REQUIRE Initial distribution $\pi_0$, observations $\mathcal{D} = \{(t_k,\vy_{k})\}_{k=1}^k$, number of particles $N$,  discretization $\{ t_j\}_{j=0}^{N_T}$
   \ENSURE Reference trajectory $\vz_{0:T}$
   \STATE Sample initial particles $\{\vx_{0}^i\}_{i=1}^N\sim \pi_0$ with weights $w_0^i {=} \frac{1}{N}$
   
  \FOR{$j=0$ {\bfseries to} $N_T$}
  \FOR{$i=1$ {\bfseries to} $N$}
    \STATE $a_{i} \sim \textbf{Categ}(\{ i\}_{i=1}^N, \vw_{j})$ \COMMENT{Draw ancestor}
    \STATE $\vx_{t_j}^i = \vx_{t_{j-1}}^{a_i}+ f(\vx_{t_{j-1}}^{a_i}, t) \diff t + \sigma \epsilon$ \COMMENT{$\epsilon \sim \mathcal{N}(0, \diff t)$}
    \STATE $\vx_{0:t_j}^i \leftarrow \{\vx_{0:t_{j-1}}^{a_i}, \vx_{t_{j}}^{i} \}$ \COMMENT{Append trajectories}
    \STATE $w_{t_j}^i \leftarrow p(\vy_{t_j} \mid \vx_{0:t_{j-1}}^i, \vy_{0:t_{j}})$ \COMMENT{Compute weights}
    \ENDFOR
  \ENDFOR
  \STATE $\vz_{0:T} \sim \textbf{Categ}(\{\vx_{0:T}^i\}_{i=1}^N, \vw_{T})$ \COMMENT{Sample reference}
\end{algorithmic}
\end{algorithm}
\begin{figure*}[t!]
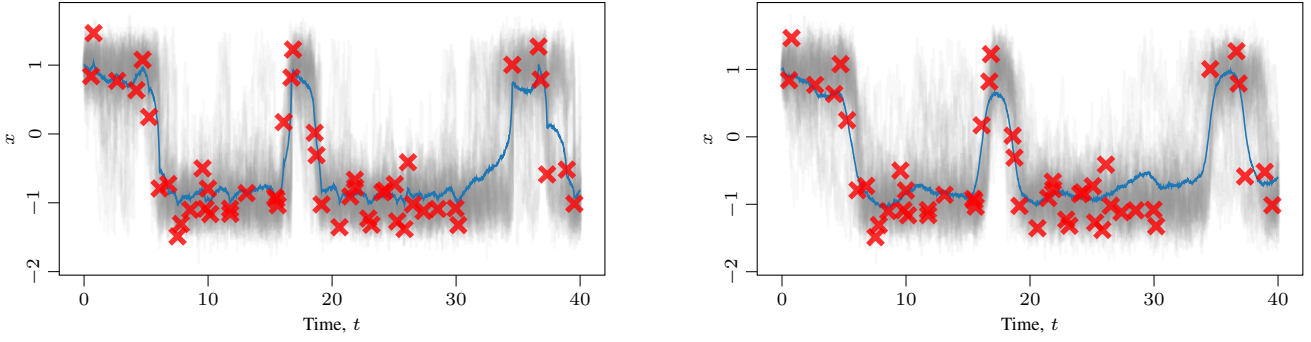

  \centering\scriptsize
  \pgfplotsset{scale only axis,y tick label style={rotate=90},xtick={0,10,20,30, 40}}
  \setlength{\figurewidth}{.4\textwidth}
  \setlength{\figureheight}{.5\figurewidth}  
  \setlength{\blockwidth}{\columnwidth}
  \begin{subfigure}[b]{\blockwidth}
    \centering
    \input{fig/dw/smoother.tex}
  \end{subfigure}
  \hfill
  \begin{subfigure}[b]{\blockwidth}
    \centering
    \input{fig/dw/model.tex}
  \end{subfigure}
   \caption{(a)~Samples from the particle smoothing distribution using the known drift of the double-well system. (b)~Trajectories generated from the learned diffusion process. The mean (in blue) and the trajectory samples (in gray) follow similar patterns, and the learned SDE roughly follows the observations without access to them at sampling time.}
   \label{fig:doublewell}
\end{figure*}
\subsection{Conditional Particle Filter with Ancestral Sampling and Smoothing}\label{sec:cpfas}
We present the details of an MCMC smoother using Conditional Particle Filter with Ancestral Sampling (CPF-AS), as defined in \cite{svensson2015nonlinear}. Instead of a forward--backward particle smoother where the particle filtering weights from the forward pass between $t=0$ to $T$ are used when accruing information from $T$ to $0$, we will generate particle smoothing trajectories via MCMC steps consisting of the CPF-AS procedure. In a single run of the CPF-AS (see \cref{alg:cpf}), the trajectories are conditioned on both the observations and a reference trajectory $\{ \vz_{t_j}\}_{j=0}^{N_T}$. At each time step, we draw an ancestor for each trajectory. For particle indices in $i \in [1, N-1]$, the probability of assigning ancestor $a^i$ are proportional to the trajectory weights $w_{t_j}^{i}$. The $N$\textsuperscript{th} particle retains information from the reference trajectory, and the probability of assigning the ancestor $a^i$ to the reference trajectory is proportional to $w_{t_j}^{a^i}f(\vz_{t_j} \mid \vx_{t_{j-1}}^{a^i})$, that is, the probability that the reference trajectory sample $\vz_t$ was generated from ancestor $a^i$. After generating $N$ particles via CPF-AS, a new reference trajectory $\{\vz_{t_j}\}_{j=0}^{N_T}$ is sampled from the weights at the terminal time, proportional to $\{w_T^i \}_{i=1}^N$. The MCMC chain is initialized as in \cref{alg:init}, with a particle filtering algorithm with ancestral sampling, but no reference trajectory. 

The CPF-AS~\cite{svensson2015nonlinear, lindsten2014particle} combined with MCMC provides an algorithm for generating trajectories from the smoothing distribution. The iterative algorithm consists of creating CPF-AS trajectories, each time conditioning on the previous reference trajectory to obtain a new reference, repeated for $M$ iterations. In \cite{svensson2015nonlinear}, an analysis of the convergence rate of the MCMC to a stationary distribution is provided, while \cite{lindsten2014particle} introduces the Conditional Particle Filtering step of the algorithm.  
\begin{algorithm}[tb]
   \caption{Conditional Particle Filter with Ancestral Sampling (CPF-AS)}
   \label{alg:cpf}
   \renewcommand{\algorithmiccomment}[1]{\hfill\textcolor{gray}{\(\triangleright\) #1}}
\begin{algorithmic}
   \REQUIRE Initial distribution $\pi_0$, observations $\mathcal{D} = \{(t_k,\vy_{k})\}_{k=1}^k$, number of filtering particles $N$, initial trajectory $\vz_{0:T}$, discretization $\{ t_j\}_{j=0}^{N_T}$.
   \ENSURE New reference trajectory $\hat{\vz}_{0:T}$ 
   \STATE Sample initial particles $\{\vx_{0}^i\}_{i=1}^N\sim \pi_0$ with weights $w_0^i = \frac{1}{N}$
   
  \FOR{$j = 0$ {\bfseries to} $N_T$}
  \STATE $a_{N}\sim \textbf{Categ}( \{ i\}_{i=1}^N, \{w_{t_j}^i p(\vz_{t_j} \mid \vx_{t_{j-1}}^i) \}_{i=1}^N) $ \COMMENT{Ancestor}
  \FOR{$i=1$ {\bfseries to} $N-1$}
    \STATE $a_{i} \sim \textbf{Categ}(\{i\}_{i=1}^N,\vw_{t_j})$ 
     \STATE $\vx_{t_j}^i = \vx_{t_{j-1}}^{a_i}+ f(\vx_{t_{j-1}}^{a_i}, t) \diff t + \sigma \epsilon$ \COMMENT{$\epsilon \sim \mathcal{N}(0, \diff t)$}
  \ENDFOR
  \FOR{$i=1$ {\bfseries} to $N$}
    \STATE $\vx_{0:t_j}^i \leftarrow \{\vx_{0:t_{j-1}}^{a_i}, \vx_{t_{j}}^{i} \}$ \COMMENT{Append trajectories}
     \STATE $w_{t_j}^i \leftarrow p(\vy_{t_j} \mid \vx_{0:t_{j-1}}^i, \vy_{0:t_{j}})$ \COMMENT{Compute weights}
     \ENDFOR
   \ENDFOR
  \STATE $j \sim \textbf{Categ}(\{ i\}_{i=1}^N, \vw_{T})$ \COMMENT{Sample reference}
\STATE $\hat{\vz}_{0:T} = \vx_{0:T}^j$ \COMMENT{Set reference trajectory }
\end{algorithmic}
\end{algorithm}

\begin{figure*}[t!]
  \centering
  \scriptsize
  \pgfplotsset{scale only axis}
  \setlength{\blockwidth}{0.18\textwidth}
  \setlength{\figurewidth}{.95\blockwidth}
  \setlength{\figureheight}{\figurewidth}
  \begin{subfigure}[b]{\blockwidth}
    \centering
\begin{tikzpicture}

\begin{axis}[
height=\figureheight,
hide x axis,
hide y axis,
tick pos=left,
width=\figurewidth,
xmin=-15, xmax=15,
ymin=-15, ymax=15
]
\addplot graphics [includegraphics cmd=\pgfimage,xmin=-19.9675324675325, xmax=18.9935064935065, ymin=-19.2857142857143, ymax=19.6753246753247] {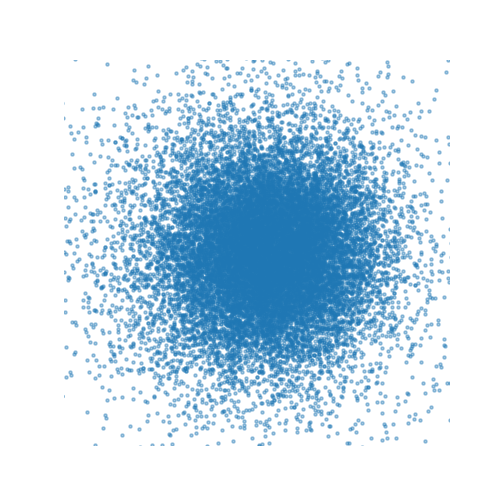};
\end{axis}

\end{tikzpicture}
  \end{subfigure}
  \hfill
  \begin{subfigure}[b]{\blockwidth}
    \centering
\begin{tikzpicture}

\begin{axis}[
height=\figureheight,
hide x axis,
hide y axis,
tick pos=left,
width=\figurewidth,
xmin=-15, xmax=15,
ymin=-15, ymax=15
]
\addplot graphics [includegraphics cmd=\pgfimage,xmin=-19.9675324675325, xmax=18.9935064935065, ymin=-19.2857142857143, ymax=19.6753246753247] {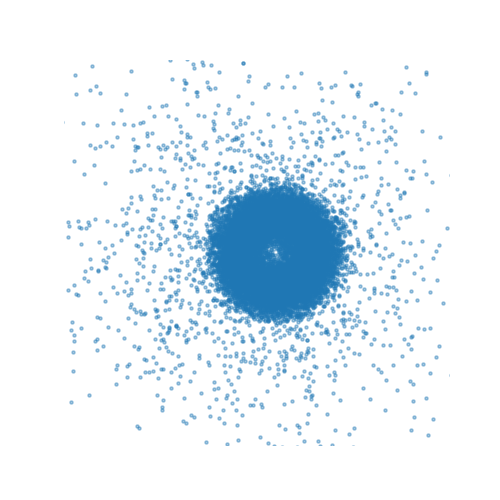};
\end{axis}

\end{tikzpicture}
  \end{subfigure}
  \hfill
  \begin{subfigure}[b]{\blockwidth}
    \centering
\begin{tikzpicture}

\begin{axis}[
height=\figureheight,
hide x axis,
hide y axis,
tick pos=left,
width=\figurewidth,
xmin=-15, xmax=15,
ymin=-15, ymax=15
]
\addplot graphics [includegraphics cmd=\pgfimage,xmin=-19.9675324675325, xmax=18.9935064935065, ymin=-19.2857142857143, ymax=19.6753246753247] {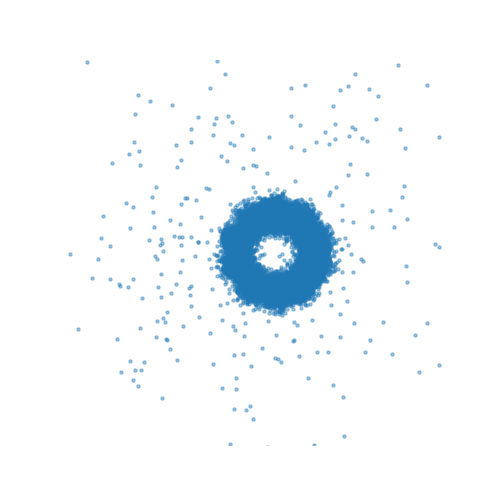};
\addplot [draw=red, fill=red!50, mark=x, only marks, mark options={solid}, mark size=4pt, opacity=0.8, line width=2pt]
table{%
x  y
1.5 3
3.26335575687742 2.42705098312484
4.35316954888546 0.927050983124842
4.35316954888546 -0.927050983124842
3.26335575687742 -2.42705098312484
1.5 -3
-0.263355756877419 -2.42705098312484
-1.35316954888546 -0.927050983124843
-1.35316954888546 0.927050983124842
-0.26335575687742 2.42705098312484
};
\end{axis}

\end{tikzpicture}
  \end{subfigure}
  \hfill
  \begin{subfigure}[b]{\blockwidth}
    \centering
\begin{tikzpicture}

\begin{axis}[
height=\figureheight,
hide x axis,
hide y axis,
tick pos=left,
width=\figurewidth,
xmin=-15, xmax=15,
ymin=-15, ymax=15
]
\addplot graphics [includegraphics cmd=\pgfimage,xmin=-19.9675324675325, xmax=18.9935064935065, ymin=-19.2857142857143, ymax=19.6753246753247] {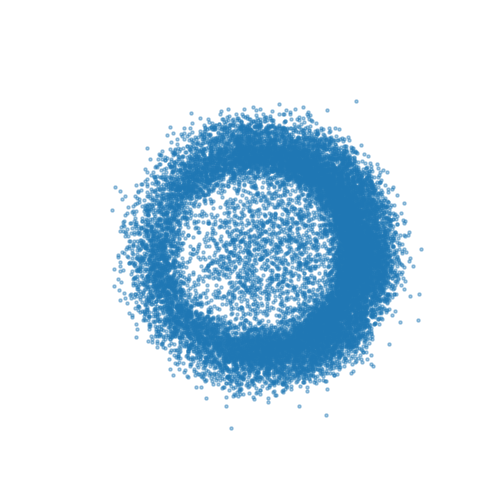};
\end{axis}

\end{tikzpicture}
  \end{subfigure}
  \hfill
  \begin{subfigure}[b]{\blockwidth}
    \centering
\begin{tikzpicture}

\begin{axis}[
height=\figureheight,
hide x axis,
hide y axis,
tick pos=left,
width=\figurewidth,
xmin=-15, xmax=15,
ymin=-15, ymax=15
]
\addplot graphics [includegraphics cmd=\pgfimage,xmin=-19.9675324675325, xmax=18.9935064935065, ymin=-19.2857142857143, ymax=19.6753246753247] {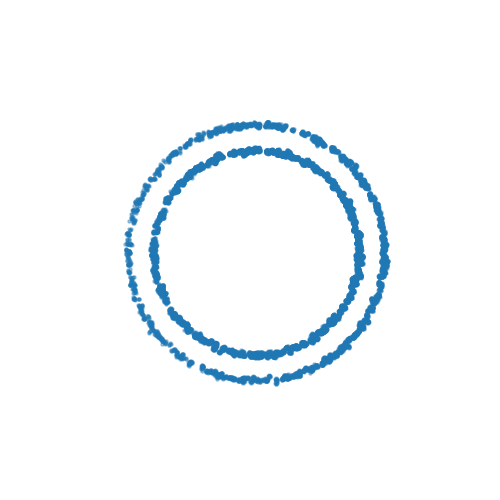};
\end{axis}

\end{tikzpicture}
  \end{subfigure}
  \begin{subfigure}[b]{\blockwidth}
    \centering
\begin{tikzpicture}

\begin{axis}[
height=\figureheight,
hide x axis,
hide y axis,
tick pos=left,
width=\figurewidth,
xmin=-15, xmax=15,
ymin=-15, ymax=15
]
\addplot graphics [includegraphics cmd=\pgfimage,xmin=-19.9675324675325, xmax=18.9935064935065, ymin=-19.2857142857143, ymax=19.6753246753247] {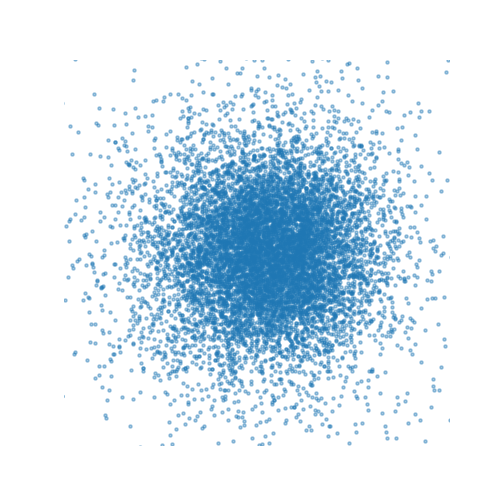};
\end{axis}

\end{tikzpicture}
  \end{subfigure}
  \hfill
  \begin{subfigure}[b]{\blockwidth}
    \centering
\begin{tikzpicture}

\begin{axis}[
height=\figureheight,
hide x axis,
hide y axis,
tick pos=left,
width=\figurewidth,
xmin=-15, xmax=15,
ymin=-15, ymax=15
]
\addplot graphics [includegraphics cmd=\pgfimage,xmin=-19.9675324675325, xmax=18.9935064935065, ymin=-19.2857142857143, ymax=19.6753246753247] {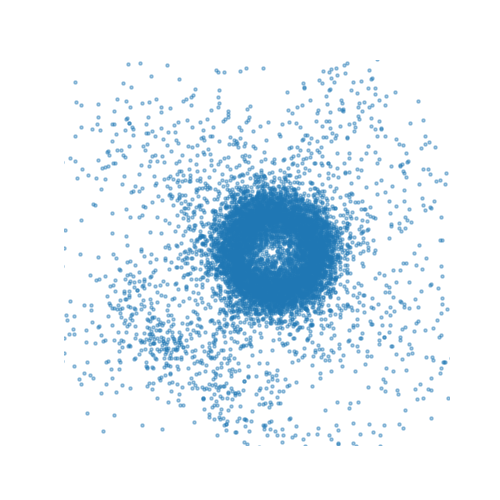};
\end{axis}

\end{tikzpicture}
  \end{subfigure}
  \hfill
  \begin{subfigure}[b]{\blockwidth}
    \centering
\begin{tikzpicture}

\begin{axis}[
height=\figureheight,
hide x axis,
hide y axis,
tick pos=left,
width=\figurewidth,
xmin=-15, xmax=15,
ymin=-15, ymax=15
]
\addplot graphics [includegraphics cmd=\pgfimage,xmin=-19.9675324675325, xmax=18.9935064935065, ymin=-19.2857142857143, ymax=19.6753246753247] {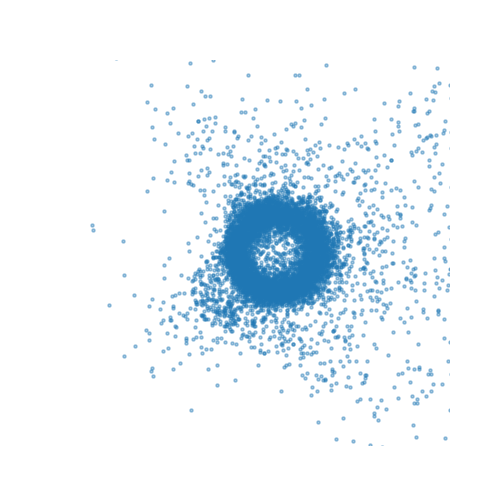};
\addplot [draw=red, fill=red!50, mark=x, only marks, mark options={solid}, mark size=4pt, opacity=0.8, line width=2pt]
table{%
x  y
1.5 3
3.26335575687742 2.42705098312484
4.35316954888546 0.927050983124842
4.35316954888546 -0.927050983124842
3.26335575687742 -2.42705098312484
1.5 -3
-0.263355756877419 -2.42705098312484
-1.35316954888546 -0.927050983124843
-1.35316954888546 0.927050983124842
-0.26335575687742 2.42705098312484
};
\end{axis}

\end{tikzpicture}
  \end{subfigure}
  \hfill
  \begin{subfigure}[b]{\blockwidth}
    \centering
\begin{tikzpicture}

\begin{axis}[
height=\figureheight,
hide x axis,
hide y axis,
tick pos=left,
width=\figurewidth,
xmin=-15, xmax=15,
ymin=-15, ymax=15
]
\addplot graphics [includegraphics cmd=\pgfimage,xmin=-19.9675324675325, xmax=18.9935064935065, ymin=-19.2857142857143, ymax=19.6753246753247] {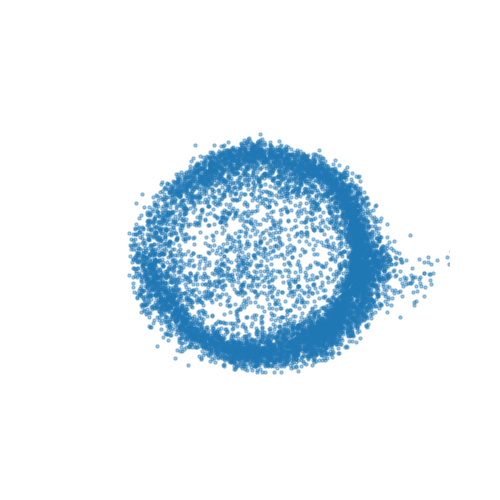};
\end{axis}

\end{tikzpicture}
  \end{subfigure}
  \hfill
  \begin{subfigure}[b]{\blockwidth}
    \centering
\begin{tikzpicture}

\begin{axis}[
height=\figureheight,
hide x axis,
hide y axis,
tick pos=left,
width=\figurewidth,
xmin=-15, xmax=15,
ymin=-15, ymax=15
]
\addplot graphics [includegraphics cmd=\pgfimage,xmin=-19.9675324675325, xmax=18.9935064935065, ymin=-19.2857142857143, ymax=19.6753246753247] {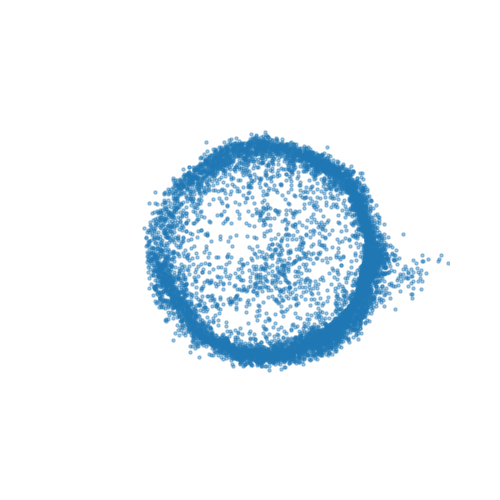};
\end{axis}

\end{tikzpicture}
  \end{subfigure}
  \caption{Particle smoother marginals  (top row) and marginals of the neural SDE model (bottom row), generated from a constrained system where the final distribution is the scikit-learn two-circles data set, and intermediate data consists of $10$ points lying on a circle with a smaller radius (in red). The marginals above are observed at times $t=\{0, 1.4, 1.5, 2.9, T\}$, where $T=3$. Without access to samples from the terminal distribution, the learned model accomplishes modeling a circle with the correct radius, but struggles to separate the two circles from each other.}
   \label{fig:scikit}
\vspace*{-1.5em}
\end{figure*}
\subsection{Efficient CPF-AS Smoothing Over a Diffusion Model}
\label{sec:algorithms}
In this section, we explain how we perform MCMC sampling over CPF-AS when multiple observations are available, and how we utilize parallelization under a setting with multi-modal observations. The extension of CPF-AS to multiple observations enables the use of the smoother for generating trajectories such that the marginal distribution matches a full observed data distribution, such as single-cell data in \cref{sec:singlecell}, instead of a single realization of a stochastic process. As explained in \cref{sec:cpfas}, the CPF-AS particle smoother generates only a single trajectory sample at each iteration. In order to reduce the heavy computational load of the particle smoother, we account for the possibly multi-modal marginals by splitting the MCMC computation to multiple, parallelizable chains. 

\paragraph{Adapted Observation Model}
In order to incorporate data from a partially observed marginal, we compute the weights of each trajectory according to a bootstrap filter proposal combined with local kernel density metric over the $H$ nearest observations to each trajectory. We may compute the log-weights as
\begin{equation}\label{eq:hnear}
\log w_{t_j}^i = -\frac{1}{2\sigma_{\mathrm{obs}}^2}\sum_{h=1}^{H} \| (\vy_{t_j,h} -\vx_{t_j}^i) \|^2,
\end{equation}
where $\vy_{t_j, h}$ is the $h\textsuperscript{th}$ nearest observation to the particle $\vx_{t_j}^i$. In practice, we apply the observation model $p(\vy_{t_j} \mid \vx_{t_j}) = \mathcal{N}(\vx_{t_j}, \sigma_{\mathrm{obs}}^2)$. 
In addition, we interpret the terminal distribution $\pi_T$ to consist of observations, so that when generating particle filtering trajectories, we observe a set of samples of size $S$, $\{ \vy_s\}_{s=1}^S$ where $\vy_s \sim \pi_T$. We may further adjust the importance of adhering to the desired terminal distribution by letting the observation model depend on time, for instance via setting $\sigma_{\mathrm{obs}}^2$ to be lower at $t=T$, or by adjusting the local kernel density estimate through modifying $H$. 

\paragraph{Parallelization}
While the CPF-AS based smoother provides accurate trajectories matching the observations, it is computationally inefficient. In order to better utilize GPU computational resources, we generate multiple MCMC chains using the CPF-AS by running multiple parallel chains for problem settings where the chains converge quickly, and the need for a large sample of trajectories arises from building a training set for the learning step detailed in \cref{sec:learning} that is sufficiently representative.
Sampling from multiple chains is further motivated by our setting, since we may observe a multimodal marginal distribution and exploration of all the modes through a single MCMC chain could prove inefficient.

\subsection{Learning Over CPF-AS Trajectories and Sampling from the Neural SDE}\label{sec:learning}
Our learning scheme consists of using the smooth trajectories obtained in the previous step to learn a dynamical system. While generating the CPF-AS smoother trajectories, we store further information to be used in the trajectory learning step. That is, while sampling from the bootstrap particle filtering proposal, we store the change in mean for each trajectory at time~$t_j$,
\begin{equation}
\vx_{t_j}^{i, \text{diff}} = (\vx_{t_{j+1}}^i + f(\vx_{t_{j+1}}^i, t_j) \Delta_j
- \vx_{t_j}^i - f(\vx_{t_{j}}^i, t_j)\Delta_j,
\end{equation}
where $\Delta_j = t_{j+1}- t_j$.
After each ancestor sampling step, we update the mean-matching trajectory history to align with the ancestors, overwriting the mean-matching history of trajectory $i$ with its ancestor's $a^i$ history. At the last step, when a new reference trajectory is sampled, we choose the mean-matching target trajectory accordingly, generating $\{\vz_{t_j}^{\text{diff}} \}_{j=1}^{N_T}$. 
We find the neural network drift $f_{\theta}$ such that the following loss is minimized for a particle $i$ at time $t_j$,
\begin{equation}\label{eq:finalloss}
\mathcal{L}(\theta, i, j) = (f_{\theta}(\vx_{t_j}^i, t_j)\Delta_j - \vz_{t_j}^{\text{diff}})^2,
\end{equation}
to learn to reverse the mean at each time-step. In practice, we apply stochastic gradient descent to a neural network $f_{\theta}$ over batches consisting of samples from the pool of particle values $\{ \vx_{t_j}^i\}_{i=1, j=1}^{N, N_T}$. After the learning step, we have access to a neural network drift $f_{\theta}$ such that the trajectories generated through Euler--Maruyama sampling of the SDE
\begin{equation}
\diff \vx_{t} = f_{\theta}(\vx_t, t)\diff t + g(t) \diff \vbeta_t, \quad \vx_0 \sim \pi_0,
\end{equation}
closely match the smoothing trajectories, and provide an approximation to its marginals without access to the observations $\{ \vy_{t_k}\}_{k=1}^K$ or samples from $\pi_T$. As such, the learned diffusion provides a scalable proxy to the CPF-AS smoother trajectories.
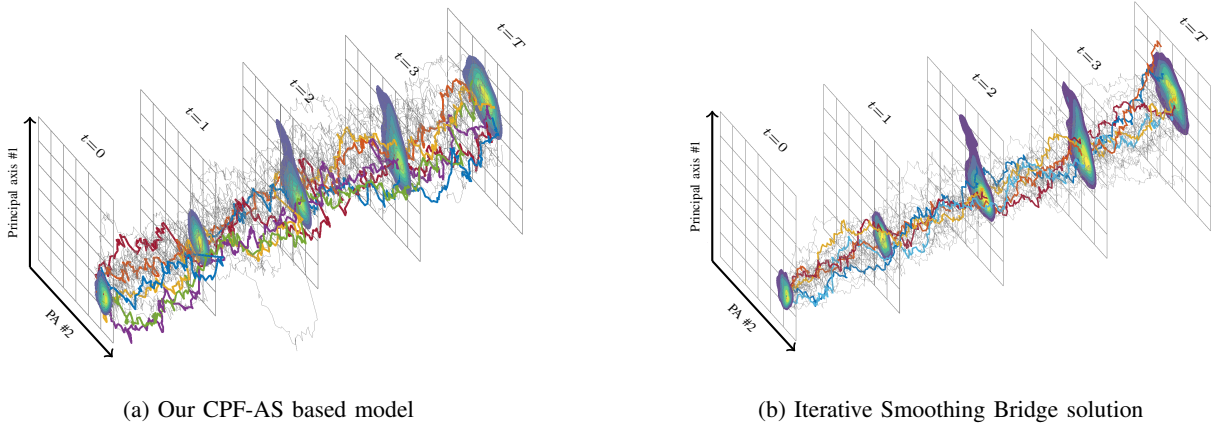
\begin{figure*}[t!]
  \centering\footnotesize
  \setlength{\figurewidth}{\textwidth}
  \setlength{\figureheight}{\figurewidth}  
  \setlength{\blockwidth}{.95\columnwidth}
  \hfill
  \begin{subfigure}[b]{\blockwidth}
    \centering
    \begin{tikzpicture}[scale=1]
        \node[anchor=south west] (I) at (-.25cm,-0.45cm) {\includegraphics[trim=80 80 80 20, clip, width=.82\textwidth]{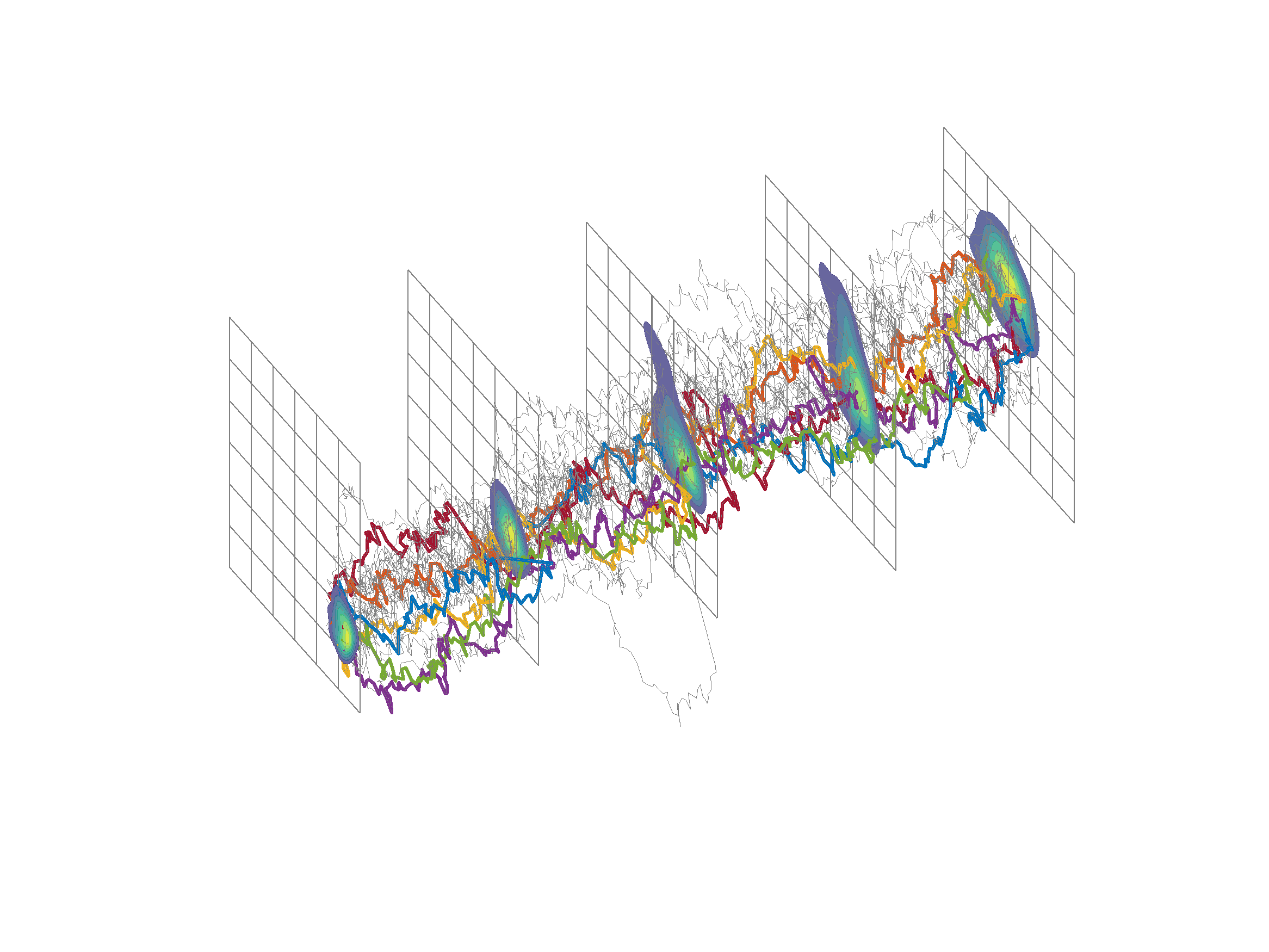}};	%

        \coordinate (a) at (.1,1.1);
        \draw[black,thick,-latex',->] (a) -- node[midway,rotate=90,yshift=6pt]{\tiny Principal axis \#1}++(0,2);
        \draw[black,thick,-latex',->] (a) -- node[midway,rotate=-45,yshift=-6pt]{\tiny PA \#2} ++(1.1,-1.2);

        \foreach \x [count=\i] in {0,1,2,3,T} {
          \node[rotate=-45] at (-.45+1.39*\i,2.35+\i*.365) {\tiny $t{=}\x$};          
        }
        
    \end{tikzpicture}
    \caption{Our CPF-AS based model}
  \end{subfigure}
  \hfill
  \begin{subfigure}[b]{\blockwidth}
    \centering  
    \begin{tikzpicture}[scale=1]
        \node[anchor=south west] (I) at (-.1cm,-0.25cm) {\includegraphics[trim=250 80 220 80, clip, width=0.8\textwidth]{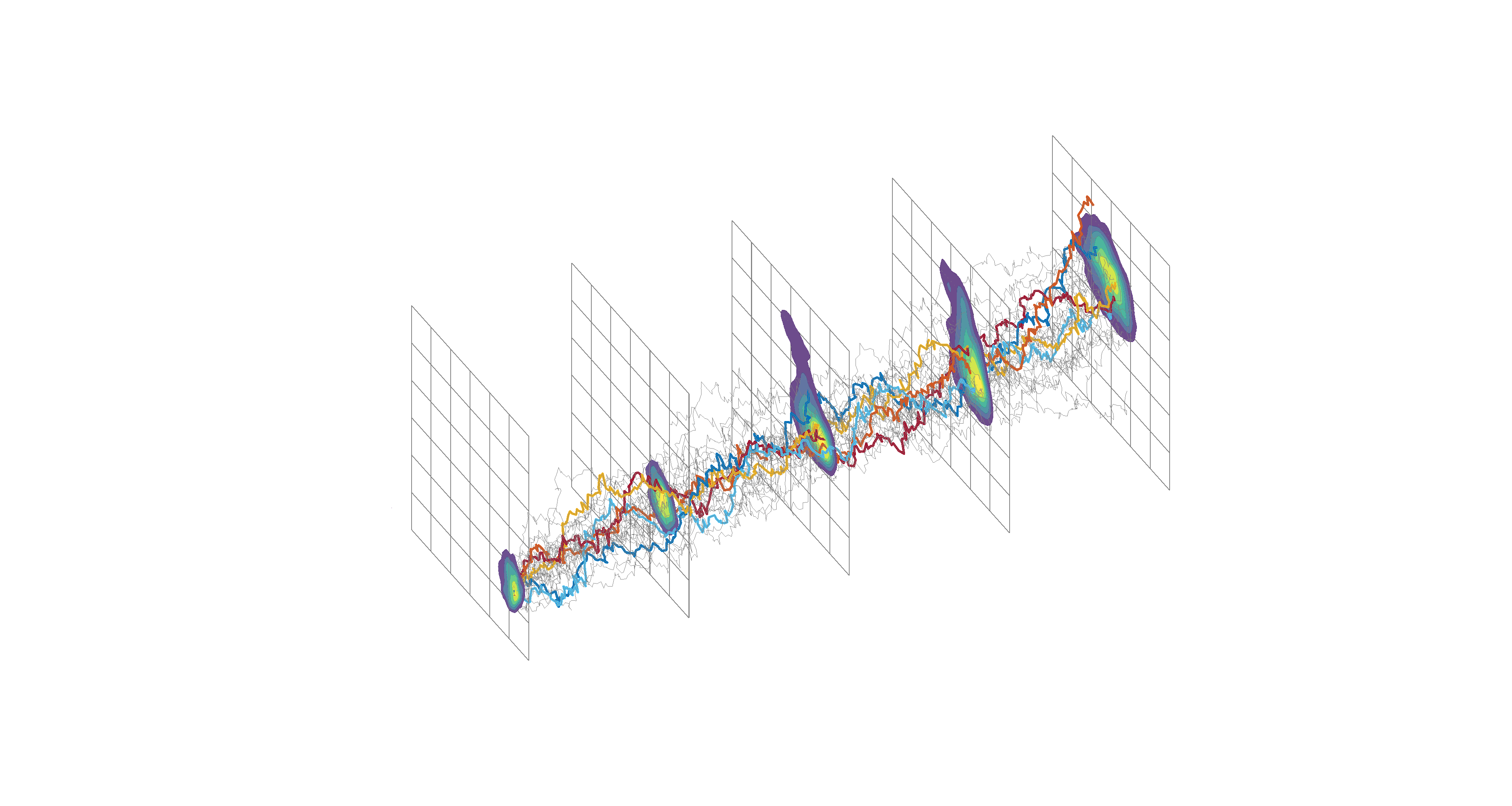}};	%

        \coordinate (a) at (.1,1.1);
        \draw[black,thick,-latex',->] (a) -- node[midway,rotate=90,yshift=6pt]{\tiny Principal axis \#1}++(0,2);
        \draw[black,thick,-latex',->] (a) -- node[midway,rotate=-45,yshift=-6pt]{\tiny PA \#2} ++(1.1,-1.2);

        \foreach \x [count=\i] in {0,1,2,3,T} {
          \node[rotate=-45] at (-.45+1.39*\i,2.35+\i*.365) {\tiny $t{=}\x$};          
        }
        
    \end{tikzpicture}\\[1em]
    \caption{Iterative Smoothing Bridge solution}    
  \end{subfigure}
  \hfill  
  \caption{The CPF-AS MCMC smoother trajectories cover the known marginals of the single-cell process (projected onto the first two principal axes for visualization on 2D planes), compared to the Iterative Smoothing Bridge which only explores high-density regions of the marginals.}
  \label{fig:singlecell}
  
\vspace*{-1.5em}
\end{figure*}

\section{Experiments}
\label{sec:experiment}
We apply the CPF-AS smoother to various settings, consisting of scenarios where the observations are from a single underlying realization of a stochastic process (\cref{sec:dw} \cref{sec:vehicle}), or a full marginal distribution (\cref{sec:scikit}, \cref{sec:singlecell}) in the middle of a transport problem. After generating the smoother trajectories, we learn the drift of a dynamical system as a neural network and compare the behaviour of the raw smoothed trajectories to the learned dynamical system.
In addition to learning a constrained generative model for time-series data, we demonstrate the capability of our method to interpolate in a setting where the sparsity of observations is artificial: we let our model utilize only every $S$\textsuperscript{th} data point, and compare the learned trajectories to the full set of observations in the \cref{sec:vehicle} interpolation experiments. \looseness-1

In all our experiments, we fix the observation model to $p(\vy_t \mid \vx_t) = \mathcal{N}(\vx_t, \sigma_{\mathrm{obs}}^2 \MI)$, allowing for scaling the observation noise at $t=T$ as explained in \cref{sec:algorithms}. We implement both the particle smoothing and trajectory reversal code in PyTorch, and utilize GPU computation to parallelize the computation over trajectories. The neural network design is as in  \cite{debortoli2021diffusion, tamir2023transport}, with four hidden layers, largest of which is 128 in width, and using a sinusoidal embedding for the time dimension.  \looseness-1

\subsection{Double-well}
\label{sec:dw}
We perform particle smoothing over a double-well system with $50$ observations. The true drift $f$ to be used within the smoothing algorithm is $f(x_t, t) = 4x_t(1-x_t^2)$,
the diffusion is $g(t) = 1$, and we set $\Delta_t=0.01$ and $T=40$. We iterate the CPF-AS particle filtering algorithm for $1000$ steps with $100$ particles used in the filter, and evaluate the performance of our method by subsequently learning the neural drift $f_{\theta}$. For the learning step, we optimize over $200$ epochs with a learning rate of $10^{-4}$ and a batch size of $2048$, and use the same process noise of $g(t) = 1$ during sampling. We use only the last $500$ reference trajectories from the smoother. The learned process manages to oscillate between the two modes of the double-well distribution matching the oscillation patterns of the particle smoother trajectories, see \cref{fig:doublewell} for a comparison. 

\subsection{2D Scikit-learn Two-circles}
\label{sec:scikit}
We perform particle smoothing and trajectory learning over a constructed data set, where the final data distribution consists of samples from the scikit-learn two circles data set, and at the middle of the process, $10$ points on a single circle are observed, and the initial distribution is a Gaussian. The purpose of this experiment is to demonstrate how CPF-AS can be used to condition on complex observation sets, and how to learn to match even such restricted generative processes. We set $\Delta_t=0.01$ and let $T=3$, and simulate from $10$ MCMC chains over $2000$ iterations of CPF-AS, each using $1000$ particles. For the training step, we set the learning rate to $10^{-4}$ and the batch size to $1024$ over $200$ epochs.
Instead of a time-invariant $g(t)$, we use a noise schedule which allows both for exploration of the state-space and is suitable for generating a sharp approximation of the scikit-learn two circles. To that end, we use a noise schedule where $g(t)=5$ at $t \in [0, T/2]$, and $g(t)$ decreases linearly to $0.01$ at $t=T$. We set the observation noise $\sigma_{\mathrm{obs}} = 0.5$ in the middle observations, scaled by a factor of $0.01$ at the scikit-learn two circles observations, and consider the $H=5$ nearest observations when computing the particle filtering weights at the terminal time, and $H=3$ at the middle. The particle smoother is able to generate trajectories which smoothly transform a Gaussian distribution first to $10$ points on a circle, and then to the scikit-learn data set. See \cref{fig:scikit} for selected marginals from the smoothing process and the learned diffusion model. \looseness-1
\begin{figure*}[t!]
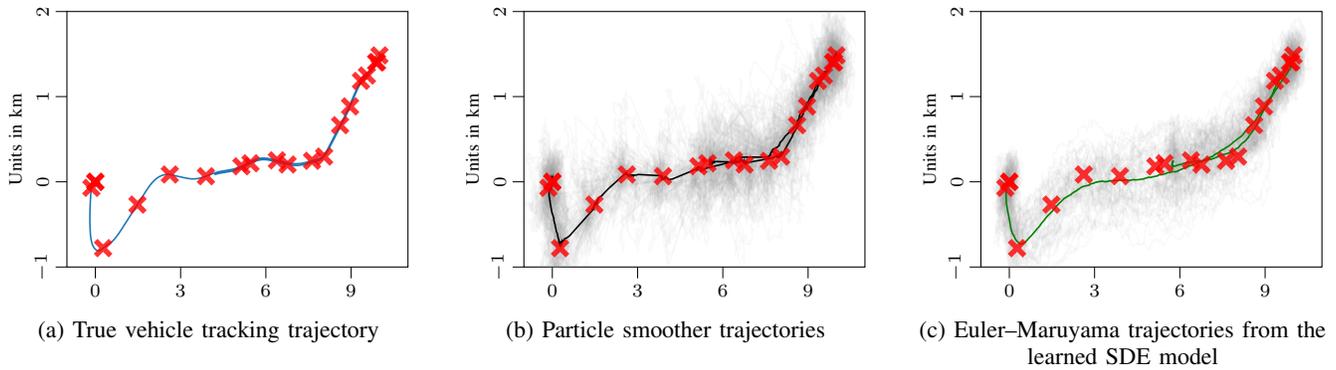

  \centering
  \scriptsize
  \def\datapath{./fig/vehicle}
  \pgfplotsset{scale only axis,y tick label style={rotate=90},xtick={ 0, 3, 6, 9},ylabel={Units in km}}
  \setlength{\blockwidth}{.33\textwidth}
  \setlength{\figurewidth}{.25\textwidth}
  \setlength{\figureheight}{.75\figurewidth}  
  \begin{subfigure}[t]{\blockwidth}
    \centering
    \input{fig/vehicle/data.tex}
    \captionsetup{justification=centering}
    \caption{True vehicle tracking trajectory}
  \end{subfigure}
  \hfill
  \begin{subfigure}[t]{\blockwidth}
    \centering
    \input{fig/vehicle/smoother.tex}
    \captionsetup{justification=centering}
    \caption{Particle smoother trajectories}
  \end{subfigure}
  \hfill
  \begin{subfigure}[t]{\blockwidth}
    \centering
    \input{fig/vehicle/model.tex}
    \captionsetup{justification=centering}
    \caption{Euler--Maruyama trajectories from the learned SDE model}
  \end{subfigure}
   \caption{A comparison of the vehicle trajectories from the data, and generated trajectories using CPF-AS particle smoothing, and learned diffusion model trajectories. The trajectory mean (bolded) follows the observations (marked by \protect\tikz[baseline=-.5ex]\protect\node[cross=5.5pt,line width=2.5pt,red]{};) in the smoother and simulated trajectories from the neural SDE, and the sampled trajectories (light gray) present similar behaviour in both plots.}
  \label{fig:vehicle}
\end{figure*}    
\subsection{Single-cell RNA Sequencing Data}
\label{sec:singlecell}
We apply the CPF-AS smoother to a single-cell RNA data set, which we preprocessed and projected into $5$ dimensions using Principal Component Analysis (PCA) as in \cite{tong2020trajectory}. Each observation represents active RNA sequence counts from an embryonic cell, which is destroyed by the measurement process, resulting in a time-series data set without any trajectory information. Instead, the observed data consists of a large number of samples from an intermediate marginal distribution at 5 points in time. We evaluate only the performance of the particle smoother in generating the marginal observations, as it accomplishes a clear improvement compared to earlier methods. 
As in earlier work \cite{tong2020trajectory, vargas2021solving}, we set the process noise to $g(t) = 1$ and discretize in time to $T=4$, $\Delta_t=0.01$. For the smoother hyperparameters, we run $8$ chains of length $500$, discarding the first half of each chain. The observation noise is set to a constant $0.3$, and the nearest 5 points are used to compute the log-weights. The CPF-AS smoother trajectories replicate the marginal distributions faithfully, see \cref{tbl:singlecell} for a comparison to earlier work using methods based on optimal transport such as \cite{tong2020trajectory} or iterative Schrödinger bridge approaches based on a reference process defined by data \cite{vargas2021solving} or using a particle filter over a diffusion model \cite{tamir2023transport}. \cref{fig:singlecell} shows how the CPF-AS MCMC smoother trajectories cover the known marginals of the single-cell process, compared to the Iterative Smoothing Bridge which only explores high-density regions of the marginals.
\begin{table}[t]
  \centering\footnotesize
  \caption{Conditional smoothing trajectories generate the lowest Earth Mover's Distance in the intermediate stages. \label{tbl:singlecell}}
  \renewcommand{\tabcolsep}{7pt}
  \begin{tabularx}{\linewidth}{l c c c c c}
    \toprule
    & \multicolumn{5}{c}{Earth mover's distance} \\
    \sc Method & \sc $t=0$ & \sc $t=1$ & \sc $t=2$ & \sc $t=3$  & \sc $t=T$ \\
    \midrule
    TrajectoryNet & $0.62$ & $1.15$ & $1.49$  & $1.26$  & $0.99$\\
    IPML & $\mathbf{0.34}$ & $1.13$ & $ 1.35$ & $1.01$ & $\mathbf{0.49}$\\
    IPFP (no obs) & $0.57$ & $1.53$ & $1.86$ & $1.32$ & $0.85$ \\
    ISB (single-cell obs) & $ 0.57$ &  $1.04$ & $1.24$ &  $0.94$ & $0.83$ \\
    CPF-AS Smoothing & $-$ & $\mathbf{0.85}$ & $\mathbf{0.93}$& $\mathbf{0.66}$ & $0.94$ \\ %
    \bottomrule
  \end{tabularx}
\end{table}
\subsection{Vehicle Tracking}
\label{sec:vehicle}
We model a 2D vehicle tracking system (a segment of data from \cite{solin15state}), where the observations give the location of a single vehicle at 1000 points in time. We sparsify the observations by allowing our model to use only every 50\textsuperscript{th} data point, with the goal of generating plausible and smooth trajectories between the observations, while comparing to the real tracking data. The observation noise of this state-space system is set to $\sigma_{\mathrm{obs}}^2 = 0.01$, the drift $f$ to zero, and the process noise to $g(t) = 0.1$, with the process starting from a single point, $\vy=(0, 0)$. We set the time step to $\Delta_t=0.01$. In order to obtain particle smoothing trajectories using CPF-AS, we ran a single MCMC chain of length $400$, and discarded the first half as burn-in. In each iteration, we ran the CPF-AS filter using 1000 particles. When optimizing the neural drift in the learning step, we applied a learning rate of $10^{-4}$ over 300 epochs of the training data set consisting of samples from the particle smoothing trajectories, with a batch size of 2048. The mean squared error (MSE) of the mean smoothing trajectory compared to the true observations was $0.215$, while the MSE of the mean over learned SDE trajectories was $0.1625$. In \cref{fig:vehicle} we compare the behaviour of the full trajectories to the means of particle smoothing trajectories and learned SDE trajectories. The latter remains a plausible approximation to the observations, without access to the sparse observations at sampling time. \looseness-1
\section{Discussion and Conclusions}
\label{sec:discussion}
The MCMC smoother iteratively applying the Conditional Particle Filter with Ancestral Sampling (CPF-AS) provides an accurate particle smoothing distribution, but is slow to generate new trajectories even after convergence. 
We have presented a new approach assimilating information from sparse observations or full marginals to the CPF-AS particle filter, including a learning step to find a neural network drift to approximate the behaviour of the smoother. The learned neural diffusion can be used to simulate trajectories under complex constraints on both the intermediate marginal distributions and the terminal distribution $\pi_T$, and it allows for an efficient approximation of the MCMC based CPF-AS smoothing trajectories. Further, our method allows for studying the system behaviour due to the analytic expression provided via the learned drift function, allowing for instance to restart the system from an intermediate point in time or to discard the observational data yet still generate novel trajectories from an approximation to the particle smoothing trajectories. 

For future work, we propose additional guidance to the MCMC smoother via learned diffusion, where a guided drift could aid the exploration of the CPF-AS trajectories, speeding up convergence. In order to assure that the full terminal distribution is explored efficiently, in the experiments in \cref{sec:experiment} we often set the process noise $g(t)$ to a high level, especially when conditioning on observations from a complex, high-variance marginal compared to more learning-based approaches such as \cite{tamir2023transport}. An adaptive drift function used within CPF-AS could plausibly be used to replace the high noise level, and to create easier to learn trajectories due to their less noisy nature. As demonstrated by our experiments, the learned SDE manages to mimic the behaviour of the particle smoothing trajectories well in settings related to learning time series, but does not generate a sharp enough replica of difficult terminal constraints, such as the scikit-learn two circles data set. Combining the trajectory learning step over smoothing trajectories with information from samples of the terminal distribution at training time could further improve the quality of the generated data under such challenging constraints. 
\section*{Acknowledgements and Disclosure of Funding}
Authors acknowledge funding from the Academy of Finland (grant 339730), and the computational resources provided by the CSC -- IT Center for Science, Finland. We wish to thank Prakhar Verma for his helpful advice on the code of \cite{verma2023variational}.

\vspace*{-0.5em}
{\small
\bibliographystyle{IEEEtran}
}
\end{document}